\documentclass[nohyperref]{article}

\usepackage[numbers]{natbib}

\usepackage{microtype}
\usepackage{graphicx}
\usepackage{subfigure}
\usepackage{booktabs} 

\usepackage{hyperref}

\usepackage[accepted]{icml2022}

\usepackage{amsmath}
\usepackage{amssymb}
\usepackage{mathtools}
\usepackage{amsthm}

\usepackage[capitalize,noabbrev]{cleveref}

\theoremstyle{plain}

\theoremstyle{definition}

\theoremstyle{remark}

\usepackage[textsize=tiny]{todonotes}

\usepackage{graphicx} 
\usepackage{amsfonts}
\usepackage{amsmath,soul}
\usepackage{amssymb}
\usepackage[capitalize]{cleveref} 
\usepackage{units}

\usepackage{dsfont} 
\usepackage{balance}
\usepackage[font=small]{caption}
\usepackage{bm} 
\usepackage{wrapfig}
\usepackage{float} 

\usepackage{color}

\def\name/{\textit{Spectral PINNs}}

\usepackage{tikz,mathtools}
\crefformat{equation}{(#2#1#3)}
\Crefformat{equation}{Equation~(#2#1#3)}
\Crefname{equation}{Equation}{Equations}

\usetikzlibrary{shapes,positioning,automata,arrows,fit,backgrounds,calc}
\tikzstyle{block} = [draw, fill=blue!20, rectangle,minimum height=1em,
minimum width=2em] \tikzstyle{sum} = [draw, fill=blue!20, circle, node
distance=1cm] \tikzstyle{input} = [coordinate] \tikzstyle{output} =
[coordinate] \tikzstyle{pinstyle} = [pin edge={to-,thin,black}]
\usetikzlibrary{trees} \usetikzlibrary{decorations.pathmorphing}
\usetikzlibrary{decorations.markings}
\definecolor{darkgreen}{rgb}{0,0.5,0}
\definecolor{darkred}{rgb}{220,20,60}

\newcommand{\cmmnt}[1]{\ignorespaces}

\newcommand{\bit}{\begin{itemize}}
\newcommand{\ei}{\end{itemize}}

\icmltitlerunning{Multiscale Neural Operators}

\begin{document}

\twocolumn[
\icmltitle{Multiscale Neural Operator: \\ Learning Fast and Grid-independent PDE Solvers}


\icmlsetsymbol{equal}{*}

\begin{icmlauthorlist}
\icmlauthor{Bj\"orn L\"utjens}{hsl}
\icmlauthor{Catherine H. Crawford}{ibm}
\icmlauthor{Campbell Watson}{ibm}
\icmlauthor{Chris Hill}{eaps}
\icmlauthor{Dava Newman}{hsl,ml}
\end{icmlauthorlist}

\icmlaffiliation{hsl}{Human Systems Laboratory, Department of Aeronautics and Astronautics, MIT}
\icmlaffiliation{ibm}{Future of Climate, IBM Research}
\icmlaffiliation{eaps}{Department of Earth, Atmospheric and Planetary Sciences, MIT}
\icmlaffiliation{ml}{MIT Media Lab}

\icmlcorrespondingauthor{Bj\"orn L\"utjens}{lutjens@mit.edu}
\icmlkeywords{Machine Learning, ICML}

\vskip 0.3in
]



\printAffiliationsAndNotice{\icmlEqualContribution} 

\begin{abstract}
Numerical simulations in climate, chemistry, or astrophysics are computationally too expensive for uncertainty quantification or parameter-exploration at high-resolution. 
Reduced-order or surrogate models are multiple orders of magnitude faster, but traditional surrogates are inflexible or inaccurate and pure machine learning (ML)-based surrogates too data-hungry. We propose a hybrid, flexible surrogate model that exploits known physics for simulating large-scale dynamics and limits learning to the hard-to-model term, which is called parametrization or closure and captures the effect of fine- onto large-scale dynamics. Leveraging neural operators, we are the first to learn grid-independent, non-local, and flexible parametrizations. Our \textit{multiscale neural operator} is motivated by a rich literature in multiscale modeling, has quasilinear runtime complexity, is more accurate or flexible than state-of-the-art parametrizations and demonstrated on the chaotic equation multiscale Lorenz96.
\end{abstract}



\section{Introduction}

\begin{figure}[t]
\centering     
		\includegraphics[width=1.\columnwidth]{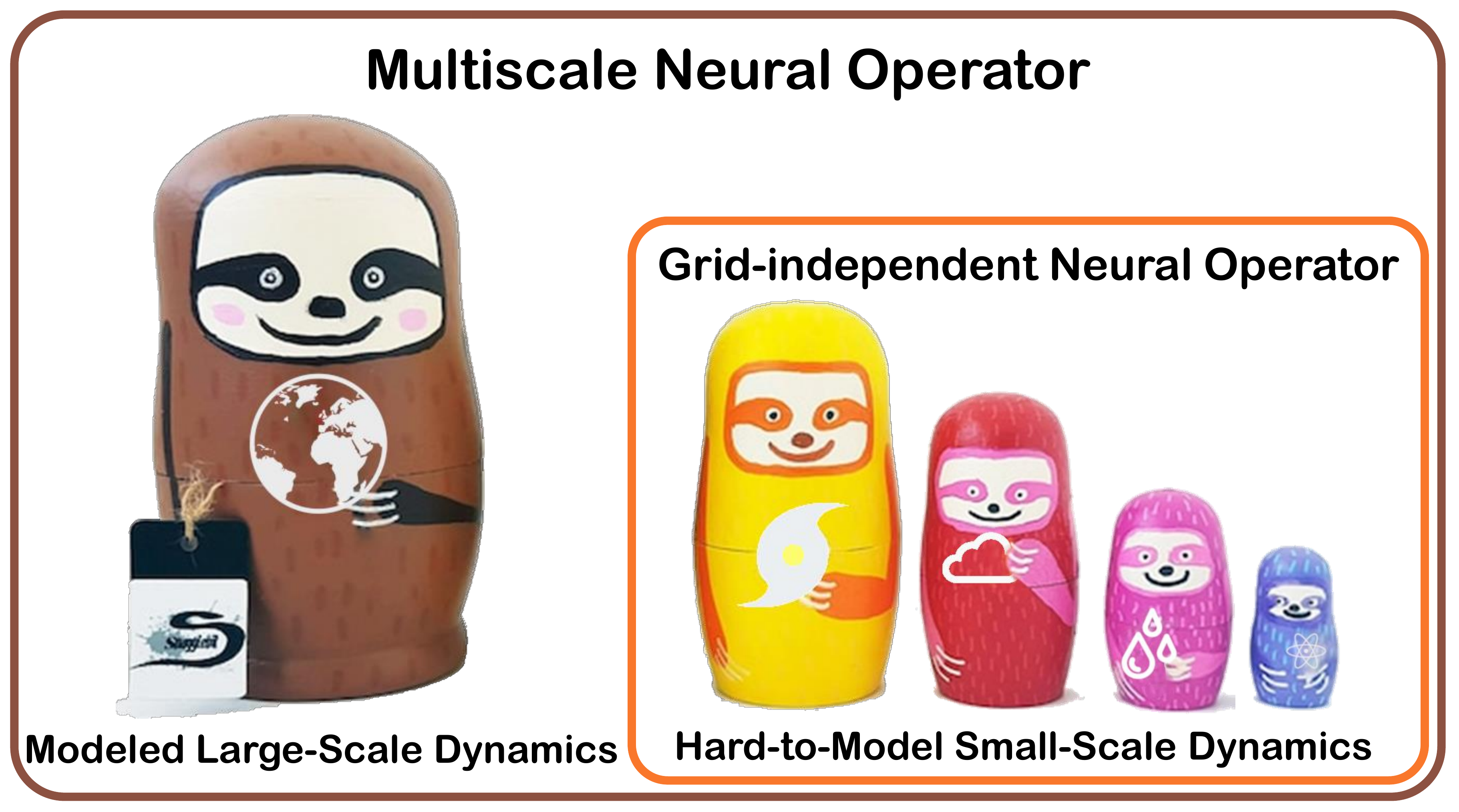}
\caption[Multiscale]{\textbf{Multiscale neural operator.} Explicitly modeling all scales of Earth's weather is too expensive for traditional and learning-based solvers~\citep{Palmer_2019}. Our multiscale neural operator dramatically reduces the computational cost by modeling the large-scale explicitly and learning the effect of fine- onto large-scale dynamics; such as turbulence slowing down a river stream. We embed grid-independent neural operators in the large-scale physical simulations as ``parametrizations``, conceptually similar to Matryoshka dolls. Image based on~\cite{Matryoshka_2022}}
\label{fig:multiscale_dolls}
\end{figure}

Climate change increases the likelihood of storms, floods, wildfires, heat waves, biodiversity loss and air pollution~\citep{IPCC_2018}. 
Decision-makers rely on climate models to understand and plan for changes in climate, but current climate models are computationally too expensive: as a result, they are hard to access, cannot predict local changes ($<10 km$), fail to resolve local extremes (e.g., rainfall), and do not reliably quantify uncertainties~\citep{Palmer_2019}. 
For example, running a global climate model at $1km$ resolution can take ten days on a $4888\times$GPU node supercomputer, consuming the same electricity as a coal power plants generates in one hour~\citep{Fuhrer_2018}. 
Similarly, in molecular dynamics~\citep{batzner22equivariantmolecule}, chemistry~\citep{behler11chem}, biology~\citep{yazdani20bioparam}, energy~\citep{zhang19power}, astrophysics or fluids~\citep{duraisamy19turbreview}, scientific progress is hindered by the computational cost of solving partial differential equations (PDEs) at high-resolution~\citep{Karniadakis_2021}. We are proposing the first PDE surrogate that quickly computes approximate solutions via correcting known large-scale simulations with learned, grid-independent, non-local parametrizations.

Surrogate models are fast, reduced-order, and lightweight copies of numerical simulations~\citep{quarteroni14roms} and of significant interest in physics-informed machine learning~\citep{kashinath21review,Reichstein_2019,Karpatne_2019,ganguly14weathermlreview}. Machine learning (ML)-based surrogates have simulated PDEs up to $1-3$ order of magnitude faster than traditional numerical solvers and are more flexible and accurate than traditional surrogate models~\citep{Karniadakis_2021}. However, pure ML-based surrogates are too data-hungry~\citep{rasp20weatherbench}; so, hybrid ML-physics models are created, for example, via incorporating known symmetries~\citep{bronstein21geometric,batzner22equivariantmolecule} or equations~\citep{willard22scimlreview}. Most hybrid models represent the solution at the highest possible resolution, which becomes computationally infeasible in multiscale or very high-resolution physics; even with optimal runtime~\citep{pavliotis08multiscale,peng21biomultiscale}. 

As depicted in~\cref{fig:multiscale_dolls,fig:model_architecture}, we simulate multiscale physics by running easy-to-acces large-scale models and focusing learning on the challenging task: \textit{How can we model the influence of fine- onto large-scale dynamics, i.e., what is the subgrid parametrization term?} The lack of accuracy in current subgrid parametrizations, also called closure or residual terms, is one of the major sources of uncertainty in multiscale systems, such as turbulence or climate~\citep{Palmer_2019,gentine18params}. Learning subgrid parametrizations can be combined with incorporating equations as soft~\citep{raissi19pinns} or hard~\citep{beucler21constraints} constraints. Various works learn subgrid parametrizations, but are either inaccurate, hard to share or inflexible because they are local~\citep{gentine18params}, grid-dependent~\citep{laypeyre19cnnparam}, or domain-specific~\citep{behler07chemparam}, respectively as detailed in~\cref{sec:params}. We are the first to formulate the parametrization problem as learning neural operators~\citep{Anandkumar_2020} to represent non-local, flexible, and grid-independent parametrizations. 

We propose, \textit{multiscale neural operator} (MNO), a novel learning-based PDE surrogate for multiscale physics with the key contributions: 
\bit
  \item A learning-based multiscale PDE surrogate that has quasilinear runtime complexity, leverages known large-scale physics, is grid-independent, flexible, and does not require autodifferentiable solvers. 
  \item The first surrogate to approximate grid-independent, non-local parametrizations via neural operators
  \item Demonstration of the surrogate on the chaotic, coupled, multiscale PDE: multiscale Lorenz96
\ei


\section{Related works}

We embed our work in the broader field of physics-informed machine learning and surrogate modeling. We propose the first surrogate that corrects a coarse-grained simulation via learned, grid-independent, non-local parameterizations.

\paragraph{Direct numerical simulation.} Despite significant progress in simulating physics numerically it remains prohibitively expensive to repeatedly solve high-dimensional partial differential equations (PDEs)~\citep{Karniadakis_2021}. For example, finite difference, element, volume, and (pseudo-)spectral methods have to be re-run for every choice of initial or boundary condition, grid, or parameters~\citep{farlow93pdes,boyd13chebyshev}. The issue arises if the chosen method does not have optimal runtime, i.e., does not scale linearly with the number of grid points, which renders it infeasibly expensive for calculating ensembles~\citep{boyd13chebyshev}. Select methods have optimal or close-to-optimal runtime, e.g., quasi-linear $O(N\log N)$, and outperform machine learning-based methods in runtime and accuracy, but their implementation often requires significant problem-specific adaptations; for example multigrid~\citep{briggs00multigrid} or spectral methods~\citep{boyd13chebyshev}. We acknowledge the existence of impressive resarch directions towards optimal and flexible non-ML solvers, such as the spectral solver called Dedalus~\citep{burns20dedalus}, but advocate to simultaneously explore easy-to-adapt ML methods to create fast, accurate, and flexible surrogate models. 

\paragraph{Surrogate modeling.} Surrogate models are approximations, lightweight copies, or reduced-order models of PDE solutions, often fit to data, and used for parameter exploration or uncertainty quantificiation~\citep{Smith_2013,quarteroni14roms}. Surrogate models via SVD/POD~\citep{chatterjee00pod}, Eigendecompositions/KLE~\citep{fukunaga70kle}, Koopman operators/DMD~\citep{williams15koopman}, take simplying assumptions to the dynamics, e.g., linearizing the equations, which can break down in high-dimensional or nonlinear regimes~\citep{quarteroni14roms}. Our work leverages the expressiveness of neural operators as universal approximations~\citep{chen95universaloperator} to learn fast high-dimensional surrogates that are accurate in nonlinear regimes~\citep{Lutjens_2021b, yuval21stable, Costa_2020,nogueira21rom}. 
\textbf{Pure ML-based} surrogate models have shown impressive sucess in approximating dynamical systems from ground-truth simulation data -- for example with neural ODEs~\citep{Rackauckas_2020,Chen_2018,hasani21ltc}, GNNs~\citep{brandstetter2022gnnpdes,cachay21graphino}, CNNs~\citep{stachenfeld2022learned}, neural operators~\citep{Li_2021,Anandkumar_2020,pathak22fourcastnet,lu21deeponet,jiang21dte}, RNNs~\citep{kani17drrnn,rasp20weatherbench}, GPs~\citep{chakraborty21gps}, reservoir computing~\citep{pathak18reservoircomputing,nogueira21rom}, or transformers~\citep{chattopadhyay20transformers} -- but, without incorporating physical knowlege become data-hungry and poor at generalization~\citep{Karniadakis_2021,beucler21ciml}.

\paragraph{Physics-informed machine learning.}
Two main approaches of incorporating physical knowledge into ML systems is via known symmetries~\citep{bronstein21geometric} or equations~\citep{Karniadakis_2021}. Our approach leverages known equations for computing a coarse-grid prior; which is complementary to using known equations as soft~\citep{raissi19pinns,lee20vaerom,yang21gans,wu20pinngans,zhang18moleculeparam,yazdani20bioparam} or hard constraints~\citep{greydanus19hamiltonian,lutter2018dln,beucler21constraints,donti2021dc3,beucler19energycons,pengzhan20sympnet} as these methods can still be used to constrain the learned parametrization. In terms of symmetry, our approach exploits translational equivariance via Fourier transformations~\citep{Li_2021}, but can be extended to other frameworks that exploit in- or equivariance of PDEs~\citep{olver86symmetry} to rotational~\citep{fuchs20se3,thomas18tensorfield}, Galilean~\citep{wu18rans,prakash21les}, scale~\citep{beucler21ciml}, translational~\citep{subel21multiscaleburgers}, reflectional~\citep{cohen17steerablecnns} or permutational~\citep{zhou20gnns} transformations.

The field of physics-informed machine learning is very broad, as reviewed most recently in~\citep{willard22scimlreview} and~\citep{Karniadakis_2021,carleo19mlps,karpatne17review}. We focus on the task of learning fast and accurate surrogate models of fine-scale models when a fast and approximate coarse-grained simulation is availabe. This task differs from other interesting research areas in equation discovery or symbolic regression~\citep{brunton16sindy,long18pdenet,long19pdenet2,liu21newphysics,qian22dcode}, downscaling or superresolution~\citep{xie18tempogan,bode21pinnsuperresolution,kurinchi21wisosuper,stengel20phiregan,vandal17deepsd,groenke20climalign}, design space exploration or data synthesis~\citep{chen20padgan,chan19geological}, controls~\citep{bieker20control} or interpretability~\citep{toms20interpret,mcgraw18granger}. Our work is complementary to data assimilation or parameter calibration~\citep{jia19pgrnn,jia21pgml,karpatne17pgnn,zhang19power,bonavita20mlassim} which fit to observational data instead of models and differs from inverse modeling and parameter estimation~\citep{parish16rans,hamilton17inverse,yin21aphinity,long18hybridnet} which fit parametrizations that are independent of the previous state. 

\paragraph{Correcting coarse-grid simulations via parametrizations.}\label{sec:params} Problems with large domains are often solved via multiscale methods~\citep{pavliotis08multiscale}. Multiscale methods simulate the dynamics on a coarse-grid and capture the effects of small-scale dynamics that occur within a grid cell via additive terms, called subgrid parametrizations, closures, or residuals~\citep{pavliotis08multiscale,Mcguffie_2005}. Existing subgrid parametrizations for many equations are still inaccurate~\citep{Webb_2015} and ML outperformed them by learning parametrizations directly from high-resolution simulations; for example in turbulence~\citep{duraisamy19turbreview}, climate~\citep{gentine18params}, chemistry~\citep{hansen13chemparam}, biology~\citep{peng21biomultiscale}, materials~\citep{liu22multiscale}, or hydrology~\citep{bennett20hydroparams}. The majority of ML-based parametrizations, however, is local~\citep{gentine18params,ogorman18convection,brenowitz18param,brenowitz20interpretparams,bretherton22coarse,yuval21stable,cachay21climart,bennett20hydroparams, hansen13chemparam,liu22multiscale,prakash21les,ling16rans,parish16rans,wu18rans,rasp20lorenz96online}, i.e., the in- and output are variables of single grid points, which assumes perfect scale separation, for example, in isotropic homogeneous turbulent flows~\citep{sagaut06les}. However, local parametrizations are inaccurate; for example in the case of anisotropic nonhomogeneous dynamics~\citep{sagaut06les,wang22nonlocal}, for correcting global error for coarse spectral discretizations~\citep{boyd13chebyshev}, or in large-scale climate models~\citep{dueben18nonlocalparams,pathak18reservoircomputing}. More recent works propose non-local parametrizations, but their formulations either rely on a fixed-resolution grid~\citep{wang22nonlocal,blakseth22resfcnns,laypeyre19cnnparam,chattopadhyay20lorenz}, an autodifferentiable solver~\citep{um20diffres,sirignano20dpm}, or are formulated for a specific domain~\citep{behler07chemparam}. A single work proposes non-local and grid-independent parametrizations~\citep{pathak20hrres}, but requires the explicit representation of a high-resolution state which is computationally infeasible for large domains, such as in climate modeling. We are the first to propose grid-independent and non-local parametrizations via neural operators to create fast and accurate surrogate models of fine-scale simulations.  

\paragraph{Neural operators for grid-independent, non-local parametrizations.}
Most current learning-based non-local parametrizations rely on FCNNs, CNNs~\citep{laypeyre19cnnparam}, or RNNs~\citep{chattopadhyay20lorenz}, which are mappings between finite-dimensional spaces and thus grid-dependent. In comparison, neural operators learn mappings in between infinite-dimensional function spaces~\citep{kovachki21universal} such as the Laplacian, Hessian, gradient, or Jacobian. Typically, neural operators lift the input into a grid-independent state such as Fourier~\citep{Li_2021}, Eigen-~\citep{Bhattacharya_2020}, graph kernel~\citep{li20graphoperator,Anandkumar_2020} or other latent~\citep{lu21deeponet} modes and learn weights in the lifted domain. We are the first to formulate neural operators for learning parametrizations.

\section{Approach}
\def\soll/{X} 
\def\sols/{Y_k} 
\def\xl/{x} 
\def\xlk/{x_k} 
\def\xs/{y} 
\def\tl/{t_x} 
\def\ts/{t_y} 
\def\al/{a_x} 
\def\as/{a_y} 
\def\dxl/{D_x} 
\def\dxs/{D_y} 
\def\fl/{f_x} 
\def\fs/{f_y} 
\begin{figure*}[t!]
 \centering
 \subfigure{
  \includegraphics[width=0.8\textwidth]{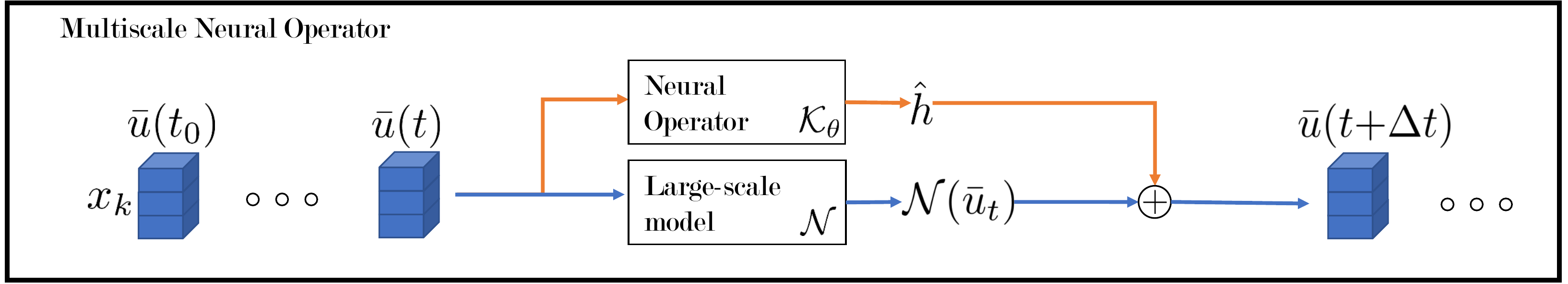}}
 \subfigure{
  \includegraphics[width=0.15\textwidth]{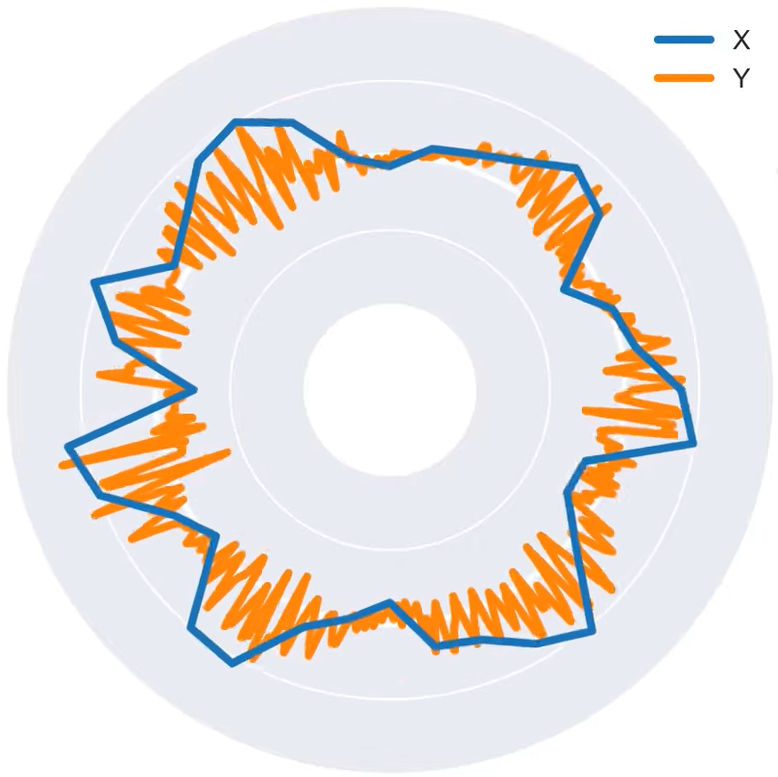}}
\caption[Superparametrizations]{Left: \textbf{Model Architecture.} A physics-based model, $\mathcal N$, can quickly propagate the state, $\bar u_t$, at a large-scale, but will accumulate the error, $h=\overline{\mathcal N (u)} - \mathcal N {\bar u}$. A neural operator, $\mathcal K_\theta$, wraps the computational and implementation complexities of unmodeled fine-scale dynamics into a non-local and grid-independent term, $\hat h$, that iteratively corrects the large-scale model. Right: \textbf{Multiscale Lorenz96.} We demonstrate multiscale neural operator (MNO) on the multiscale Lorenz96 equation, a model for chaotic atmospheric dynamics. Image:~\citep{rasp20lorenz96online}}  \label{fig:model_architecture}
\end{figure*}

We propose \textit{multiscale neural operator} (MNO): a surrogate model with quasilinear runtime complexity that exploits know coarse-grained simulations and learns a grid-independent, non-local parametrization. 
\subsection{Multiscale neural operator}\label{sec:mno}

\paragraph{Partial differential equations.}
We focus on partial differential equations (PDEs) that can be written as initial value problem (IVP) via the method of lines~\citep{schiesser91methodlines}. The PDEs in focus have one temporal dimension, $t\in[0,T]=:D_t$, and (multiple) spatial dimensions, $x=[x_1, ..., x_d]^T \in D_x$, and can be written in the iterative, explicit, symbolic form~\citep{farlow93pdes}:

\begin{equation}
\begin{aligned}
\frac{\partial u}{\partial t} - \mathcal N (u) = 0 \text{ with } &t,x\in [0,T]\times D_x\\
u(0, x) = u^0(x) , \;\mathcal B[u](t,x) = 0 \text{ with } &x\in D_x,\; \\
& (t,x){\in} [0,T]{\times} \partial D_x
\end{aligned}
\label{eq:symbolicpde}
\end{equation}

In our case, the (non-)linear operator, $\mathcal N$, encodes the \textbf{known} physical equations; for example a combination of Laplacian, integral, differential, etc. operators. Further, $u: D_t \times D_x \rightarrow D_u$ is the solution to the initial values, $u^0: D_x \rightarrow D_u$, and Dirichlet, $\mathcal B_D[u] = u - b_D$, or Neumann boundary conditions, $\mathcal B_N[u] = n^T \delta_x u - b_N$, with outward facing normal on the boundary, $n\bot \delta B$.

\paragraph{Scale separation.}
We transfer a concept from the rich and mathematical literature in multiscale modeling~\citep{pavliotis08multiscale} to consider a filter kernel operator, $\mathcal G\ast $, that creates the large-scale solution, $\bar u(x) = u(x) + u^\prime(x)$, where $u^\prime$ are the small-scale deviations and $\bar \cdot$ denotes the filtered variable, $\bar \phi (x) = \mathcal G\ast \phi = \int_{D_x} G(x,x^\prime) \phi(x^\prime) dx^\prime$. Assuming the kernel, $G$, preserves constant fields, $\bar a=a$, commutes with differentiation, $[\mathcal G\ast, \frac{\delta}{\delta s}], s=x,t$, is linear, $\overline{\phi + \psi} = \bar\phi + \bar\psi$~\citep{sagaut06les}, we can rewrite~\cref{eq:symbolicpde} to: 

\begin{equation}
\begin{aligned}
\mathcal G\ast \frac{\delta u}{\delta t} = \frac{\delta \bar u}{\delta t} &= \mathcal G \ast \mathcal N (u) \\
&= \mathcal N (\bar u) + [\mathcal G \ast, \mathcal N](u)
\end{aligned}
\label{eq:scalesep}
\end{equation}

where $[\mathcal G\ast, \mathcal N](u) =\mathcal G\ast \mathcal N (u ) - \mathcal N (\mathcal G\ast u )$ is the filter subgrid parametrization, closure term, or commutation error, i.e., the error introduced through propagating the coarse-grained solution. 

Approximations of the subgrid parametrization as an operator that acts on $\bar u$ require significant domain expertise and are derived on a problem-specific basis. In the case of isotropic homogeneous turbulence, for example, the subgrid parametrization can be approximated as the spatial derivative of the subgrid stress tensor, $[\mathcal G\ast,\mathcal N](\bar u)_\text{turbulence}\approx\frac{\delta \tau_{ij}}{\delta x_j} = \frac{\delta \overline{u_i^\prime u_j^\prime}}{\delta x_j}$~\citep{sagaut06les}. Many works approximate the subgrid stress tensor with physics-informed ML~\citep{prakash21les,ling16rans,parish16rans,wu18rans}, but are domain-specific, local, or require a differentiable solver or fixed-grid. We propose a general purpose method to approximating the subgrid parametrization, independent of the grid, domain, isotropy, and underlying solver.

\paragraph{Multiscale neural operator.}\label{sec:approach_mno}
We aim to approximate the filter commutation error, $[\mathcal G\ast, \mathcal N]\approx: h$, via learning a neural operator on high-resolution training data. Let $\mathcal K_\theta$ be a neural operator that approximates the commutation error:
\begin{equation}
\begin{aligned}
[\mathcal G\ast, \mathcal N] \approx \mathcal K_\theta: \bar U(D_x;\mathbb R^{d_u}) \rightarrow H(D_x;\mathbb R^{d_u})
\end{aligned}
\label{eq:no}
\end{equation}
where $\theta$ are the learned parameters and $\bar U, H$ are separable Banach spaces of all continuous functions taking values, $\mathbb R^{d_u}$, defined on the bounded, open set, $D_x \subset \mathbb R^{d_x}$, with norm $\lvert\lvert f\rvert\rvert_{\bar U} = \lvert\lvert f\rvert\rvert_{H} = \max_{x\in D_x} \lvert f(x) \rvert$. We embed the neural operator as an autoregressive model with fixed time-discretization, $\Delta t$, such that the final \textit{multiscale neural operator} (MNO) model is:

\begin{equation}
\begin{aligned}
\bar u(t + \Delta t) &= f(t, \bar u, \frac{\delta \bar u}{\delta x}, \frac{\delta^2\bar u}{\delta x^2},\dots) + \mathcal K_\theta(\bar u)\end{aligned}
\label{eq:mno}
\end{equation}

where $f(t, \bar u, \frac{\delta \bar u}{\delta x}, \frac{\delta^2\bar u}{\delta x^2}) = \int_t^{t+\Delta t}\mathcal N(\bar u)d\tau$ is the known large-scale tendency, i.e. one-step solution. MNO is fit using MSE with the loss function:

\begin{equation}
\begin{aligned}
L &= \mathds E_t \mathds E_{\bar u\rvert u(t)\sim p(t)} \left( \mathcal L (\mathcal K_\theta (\bar u (t)), [\mathcal G\ast, \mathcal N](u(t))\right) \\
\end{aligned}
\label{eq:loss}
\end{equation}
where the ground-truth data, $u(t)\sim p(t)$, is generated by integrating a high-resolution simulation with varying parameters, initial or boundary conditions and uniformly sampling time snippets according to the distribution $p(t)$. Similar to problems in superresolution, there exist multiple realizations of the learned commutation error, $[\mathcal G\ast, \mathcal N](\bar u)$, for a given ground-truth, $[\mathcal G\ast, \mathcal N](u)$; using MSE will learn a smooth average and future work will explore adversarial losses~\citep{goodfellow14gans} or an intersection between neural operators and normalizing flows~\citep{rezende15normflows} or diffusion-based models~\citep{dickstein15diff} to account for the stochasticity~\citep{wilks05lorenz96stochastic}. During training, the model input is generated via $\bar u(t) = \mathcal G\ast(u(t))$ and the target via 
\begin{equation}
\begin{aligned}
h_\text{target} &= \overline{\mathcal N(u)} - \mathcal N (\bar u).
\end{aligned}
\label{eq:target}
\end{equation}
During inference MNO is initialized with a large-scale state and integrates the dynamics in time via coupling the neural operator and a large-scale simulation. 

Our approach does not need access to the high-resolution simulator or equations; it only requires a precomputed high-resolution dataset, which are increasingly available~\citep{herbach20era5,jh22turbulencedata}, and allows the user to incorporate existing easy-to-access solvers of large-scale equations. There is no requirement for the large-scale solver to be autodifferentiable which significantly simplifies the implementation for large-scale models, such as in climate. If desired, our loss function can easily be augmented with a physics-informed loss~\citep{raissi19pinns} on the large-scale dynamics or parametrization term.  

\paragraph{Choice of neural operator.}
Our formulation is general enough to allow the use of many operators, such as Fourier~\citep{Li_2021}, PCA-based~\citep{Bhattacharya_2020}, low-rank~\citep{khoo19switchnet}, Graph~\citep{li20graphoperator} operators, or DeepOnet~\citep{wang21deeponet,lu21deeponet}. Because DeepONet~\citep{lu21deeponet} focuses on interpolation and assumes fixed-grid sensor data, we decided to modify Fourier Neural Operator (FNO)~\citep{Li_2021} for our purpose. FNO is a universal approximator of nonlinear operators~\citep{kovachki21universal,chen95universaloperator},   grid-independent and can be formulated as autoregressive model~\citep{Li_2021}. As there exist significant knowledge on symmetries and conservation properties of the commutation error~\citep{sagaut06les}, MNO's explicit formulation increases interpretability and ease of incorporating symmetries and constraints. With FNO, we exploit approximate translational symmetries in the data and leave novel opportunities for neural operators that exploit the full range of known equi- and invariances of the subgrid parametrization term, such as Galilean invariance~\citep{prakash21les}, for future work.

\subsection{Illustration of MNO via multiscale Lorenz96}\label{sec:gt_lorenz96}
We illustrate the idea of MNO on a canonical model of atmospheric dynamics, the multiscale Lorenz96 equation~\cite{lorenz96lorenz,thornes17lorenz96}. This PDE is multiscale, chaotic, time-continuous, space-discretized, 2D (space+time), nonlinear, displayed in~\cref{fig:model_architecture}-right and detailed in Appendix A.3. 
Most importantly, the large- and small-scale solutions, $X_k\in\mathbb R, Y_{j,k}\in\mathbb R\;\forall\;j\in\{0,...,J\},k\in\{0,...,K\}$, demonstrate the \textit{curse of dimensionality}: the number of the small-scale states grows exponentially with scale and explicit modeling becomes computationally expensive, for example, quadratic for two-scales: $O(N^2)=O(JK)$. The PDE writes:
\begin{equation}
\begin{aligned}
\frac{\delta X_k}{\delta t} &{=} X_{k-1}(X_{k+1}{-}X_{k-2}){-}X_k {+} F {-}\frac{h_s c}{b} \sum_{j=0}^{J-1}{ Y_{j,k}(X_k)}, \\
\frac{\delta Y_{j,k}}{\delta t} &{=} {-}cb Y_{j+1,k}(Y_{j+2,k}{-}Y_{j-1,k}){-}c Y_{j,k}{+}\frac{h_s c}{b}X_k.
\end{aligned}
\label{eq:lorenz96}
\end{equation}
where $F$ is the forcing, $h_s$ the coupling strength, $b$ the relative magnitude of scales, and $c$ the evolution speed. 
With the multiscale framework from~\cref{sec:mno}, we define:
\begin{equation*}
\begin{aligned}
&u(x) = [X_0, Y_{0,0}, Y_{1,0}, ..., Y_{J,0}, X_1, Y_{0,1}, ...\\
&\;\;\;\;\;\;\;\;\;\;\;, X_K, ..., Y_{J,K}]_x\;\forall x{\in} D_x{ = }\{0,...,K(J+1)\} \\
&\mathcal N(u)(x) = 
\begin{cases}
  \frac{\delta X_k}{\delta t} &\text{ if }x {=} k(J{+}1)\;\forall k{\in}\{0,\dots,K\}\\
  \frac{\delta Y_{j,k}}{\delta t} &\text{ otherwise,}
\end{cases}\\
&G(x,x^\prime) = 
\begin{cases}
  1 \text{ if }x^\prime = k(J+1)\;\forall k\in\{0,\dots,K\}\\
  0 \text{ otherwise},
\end{cases}
\end{aligned}
\label{eq:multiscale_lorenz96}
\end{equation*}
with the solution, $u$, operator, $\mathcal N$, and kernel, $G$. 

MNO learns the parametrization term via a neural operator, $\mathcal K_\theta = \hat h \approx h$, and then models:
\begin{equation}
\begin{aligned}
\frac{\delta  \hat X_k}{\delta t} &= \frac{\delta \overline {\hat  X}_k}{\delta t} + \mathcal K_\theta( \hat X_{0:K}) (k)
\end{aligned}
\label{eq:lorenz96_nn}
\end{equation}

where the known large-scale dynamics are abbreviated with $\frac{\delta \overline {\hat X_k}}{\delta t}= \hat X_{k-1}( \hat X_{k+1}- \hat X_{k-2})- \hat X_k + F$ and ground-truth parametrization is $h(x) = \{
-\frac{h_s c}{b}\sum_{j=0}^{J-1}Y_{j,k}(X_k) \text{ if } x = k(J+1)\;\forall k\in\{0,\dots, K\} \text{ and } 
0 \text{ otherwise}\}$. See Appendix A.4 for all terms. 

The parametrization, $\mathcal K_\theta$, accepts inputs that are sampled anywhere inside the spatial domain, which differs from previous local~\citep{rasp20lorenz96online} or grid-dependent~\citep{chattopadhyay20lorenz} Lorenz96 parametrizations. 

We create the ground-truth data via randomly sampled initial conditions, periodic boundary conditions, and integrating the coupled equation with a 4th-order Runge-Kutta solver. After a Lyapunov timescale the state is independent of initial conditions and we extract $4$K snippets with $T/\Delta t=400$steps length for 1-step training. This model is run autoregressively on $1$K test samples of length $T/\Delta t=400$steps, which correspond to 10 Earth days, as detailed in Appendix A.3.
\begin{figure}[t]
 \centering
  \includegraphics[width=1.\columnwidth]{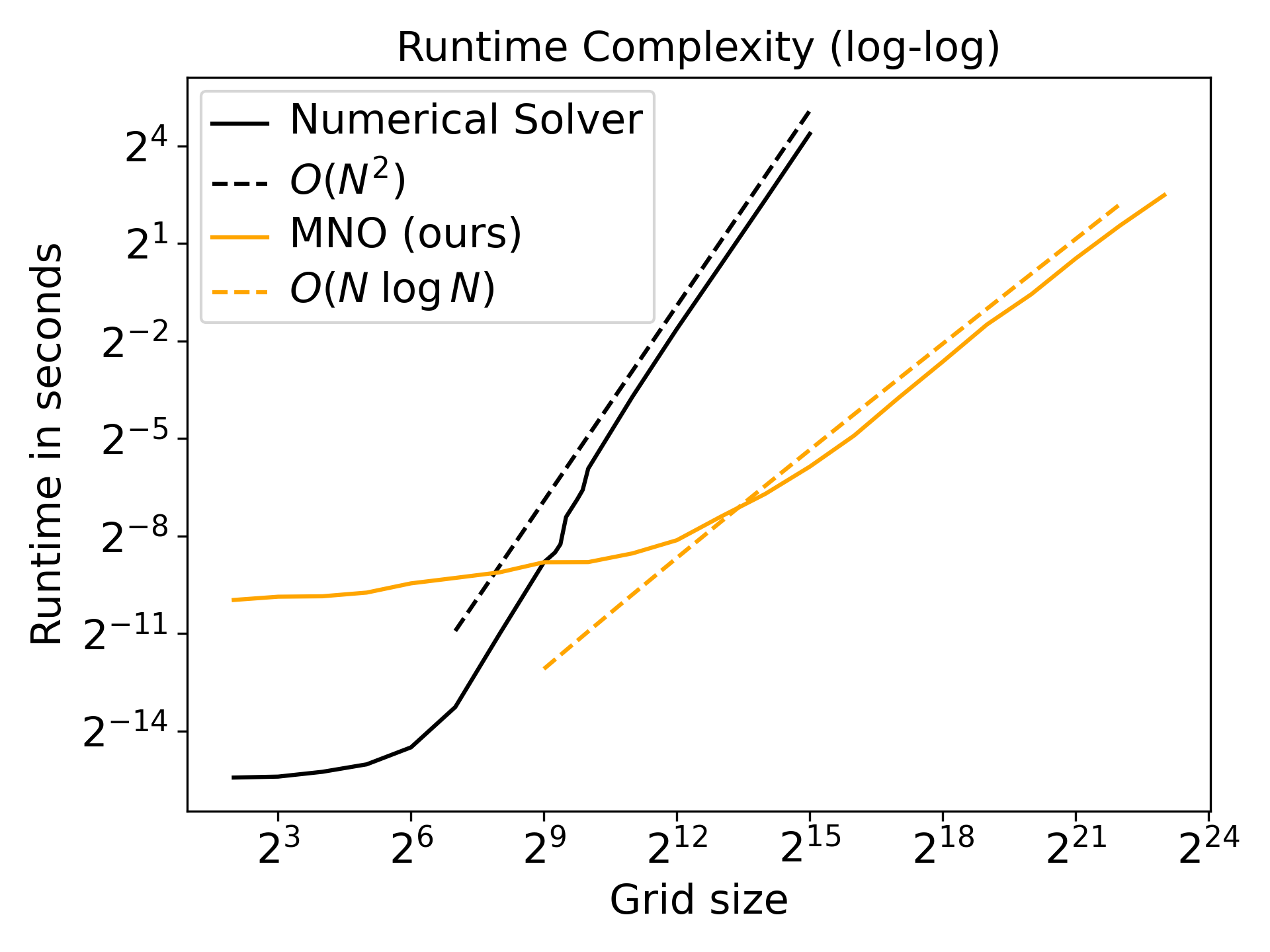}
\caption[Runtime]{\textbf{MNO is faster than direct numerical simulation.} Our proposed multiscale neural operator (orange) can propagate multiscale PDE dynamics in quasilinear complexity, $O(N\;\log N)$. For a grid with $K=2^{15}$, MNO is $\sim1000$-times faster than direct numerical simulation (black) which has quadratic complexity, $O(N^2)$}
\label{fig:runtimes_plot}
\end{figure}

\section{Results}
Our results demonstrate that multiscale neural operator (MNO) is faster than direct numerical simulation, generates stable solutions, and is more accurate than current parametrizations. We now proceed to discussing each of these in more detail.

\subsection{Runtime Complexity: MNO is faster than traditional PDE solvers}

MNO (orange in~\cref{fig:runtimes_plot}) has quasilinear, $O(N\; \log N)$, runtime complexity in the number of large-scale grid points, $N{=}K$, in the multiscale Lorenz96 equation. The runtime is dominated by a lifting operation, here a fast Fourier transform (FFT), which is necessary to learn spatial correlations in a grid-independent space. In comparison, the direct numerical simulation (black) has quadratic runtime complexity, $O(N^2)$, because of the explicit representation of $N^2{=}JK$ small-scale states. Both models are linear in time, $O(T)$. Local parametrizations can achieve optimal runtime, $O(N)$, but it is an open question if there exists a decomposition that replaces FFT to yield an optimal, non-local, grid-independent model.

We ran MNO up to a resolution of $K=2^{24}$, which would equal $75cm/px$ in a global 1D (space) climate model and only took $\approx2s$ on a single CPU. MNO is three orders of magnitude ($1000$-times) faster than DNS, at a resolution of $K=2^{15}$ or $200m/px$. 
For 2D or 3D simulations the gains of using MNO vs. DNS are even higher with $O(N^2\;\log N)$ vs. $O(N^4)$ and $O(N^3\;\log N)$ vs. $O(N^6)$, respectively~\citep{Khairoutdinov_2005}. 

The runtimes have been calculated by choosing the best of 1-100k runs depending on grid size on a single-threaded Intel Xeon Gold 6248 CPU@2.50GHz with 164Gb RAM. We time a one step update which, for DNS, is the calculation of~\cref{eq:lorenz96} and for MNO the calculation of~\cref{eq:lorenz96_nn}, i.e., the sum of a large-scale step and a pass through the neural operator. 

In~\cref{fig:runtimes_plot}, MNO and DNS plateau at low-resolution ($K<2^9$), because runtime measurement is dominated by grid-independent operations. DNS plateaus at a lower runtime, because MNO contains several fixed-cost matrix transformations. The runtime of DNS has a slight discontinuity at $K\approx 2^9$ due to extending from cache to RAM memory. We focus on a runtime comparison, but MNO also has significant savings in memory: representing the state at $K=2^{17}$ in double precision occupies $64$GB RAM for DNS and $0.5$MB for MNO. 


\subsection{MNO is more accurate than traditional parametrizations}

\begin{figure}[t]
 \centering
  \subfigure{
  \includegraphics[width=0.9\columnwidth]{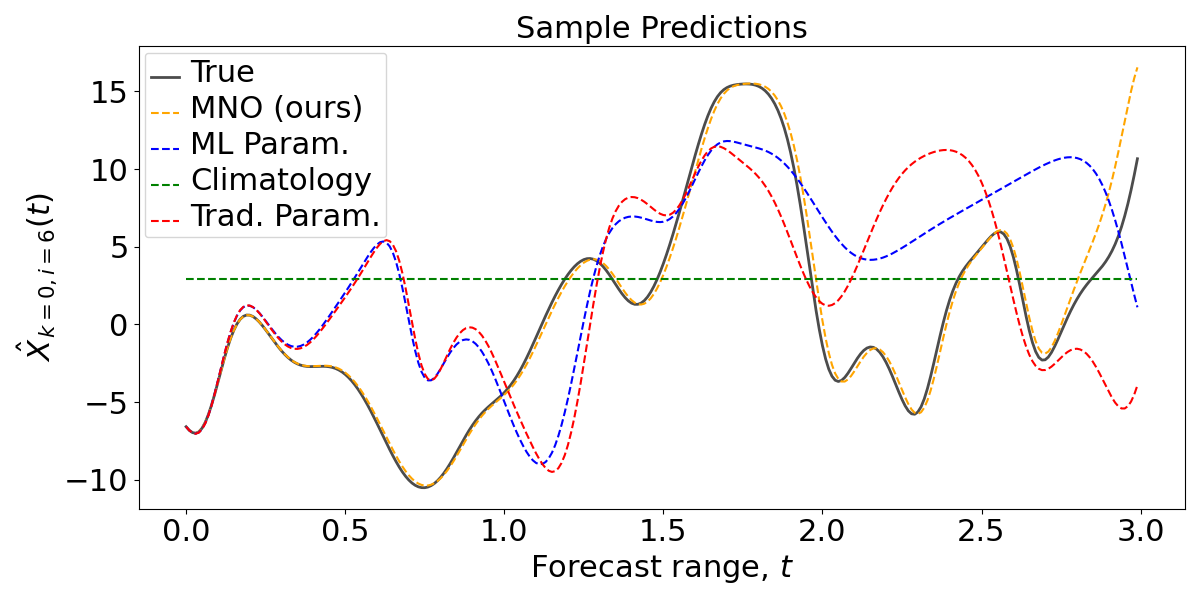}}
  \subfigure{
  \centering
  \label{tab:accuracy}
   \begin{tabular}{ll}
    \toprule
    Method     & RMSE     \\ 
    \midrule
    Climatology & $6.902$ \\
    Traditional parametrizations & $2.326$ \\
    ML-based parametrization~\cite{Rasp_2018} & $2.053$ \\
    \textbf{MNO (ours)} & $0.5067$ \\ 
    \bottomrule
  \end{tabular}
 }
\caption[Accuracy]{\textbf{Left: MNO is more accurate than traditional parametrizations.} A sample plot shows, that our proposed multiscale neural operator (yellow/orange-dotted) can accurately forecast the large-scale physics (black-solid), $X_{k=0}(t)$. In comparison, ML-based blue-dotted) and traditional (red-dotted) parametrizations quickly start to diverge. Note that the system is chaotic and small deviations are rapidly amplified; even inserting the exact parametrizations in float32 instead of float64 quickly diverges.
\textbf{Right: Accuracy.} MNO is more accurate than traditional parametrizations as measured by the root mean-square error (RMSE).
}
\label{fig:accuracy}
\end{figure}

\Cref{fig:accuracy}-left shows a forecasted trajectory of a sample at the left boundary, $k=0$, where MNO (orange-dotted) accurately forecasts the large-scale dynamics, $X_{0}(t)$, (black-solid) while current ML-based (blue-dotted)~\citep{gentine18params} and traditional parametrizations (red-dotted) quickly diverge. The quantitive comparison of RMSE and a mean/std plot~\cref{fig:mean_accuracy} over $1K$ samples and $200$steps or $10\text{days}$ ($\Delta t=0.005=36\text{min}$) confirms that MNO is the most accurate in comparison to ML-based parametrizations, 
traditional parametrizations, and a mean forecast (climatology). Note, the difficulty of the task: when forecasting \textit{chaotic} dynamics even numerical errors rapidly amplify~\citep{sagaut06les}.

\textbf{ML-based parametrizations} is a state-of-the-art (SoA) model in learning parametrizations and trains a ResNet to forecast a local, grid-independent parametrization, $h_k = \text{NN}(X_k)$, similar to~\citep{gentine18params}. The \textbf{traditional parametrizations} (trad. param.) are often used in practice and use linear regression to learn a local, grid-independent parametrization~\citep{Mcguffie_2005}. It was suggested that multiscale Lorenz96 is too easy as a test-case for comparing offline models because traditional parametrizations already perform well~\citep{rasp19lorenztooeasy}, but the significant difference between MNO and Trad. Params. during online evaluation suggests otherwise. 
The \textbf{climatology} forecasts the mean of the training dataset, $X_k(t) = 1/T \sum_{t=0}^T 1/N \sum_{i=0}^N X_{k,i}(t)$. The full list of hyperparameters and model parameters can be found in Appendix A.5.2. 
For fairness, we only compare against grid-independent methods that do not require an autodifferentiable solver; models with soft or hard constraints, e.g., PINNs~\citep{raissi19pinns} or DC3~\citep{donti2021dc3}, are complementary to MNO. 

\subsection{MNO is stable}
\begin{figure}[t]
 \centering
  \subfigure[Long-term sample forecast, $X_0(t)$]{
  \includegraphics[width=1.\columnwidth]{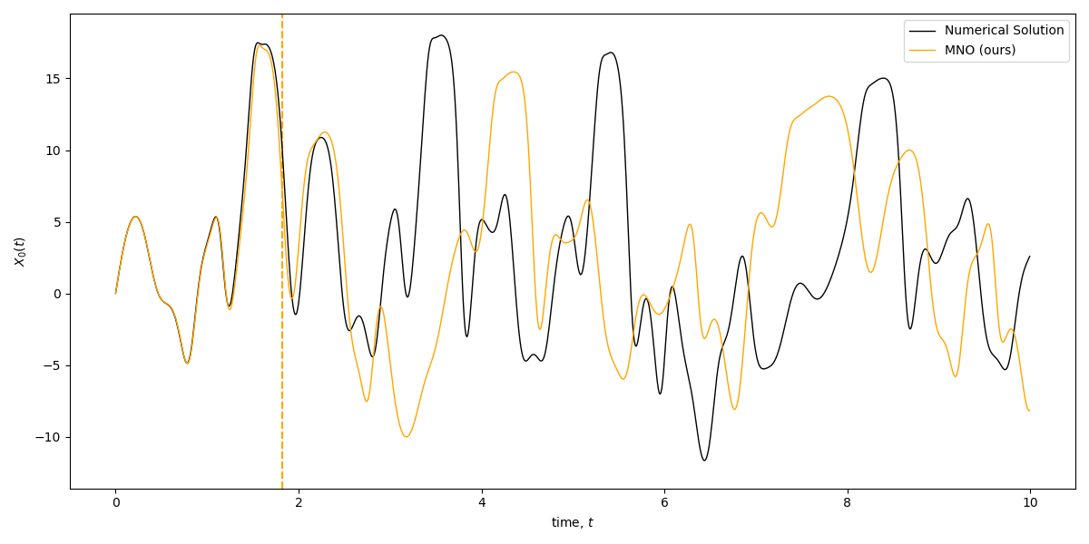}}
  \\
  \subfigure[Error over time]{
  \includegraphics[width=1.\columnwidth]{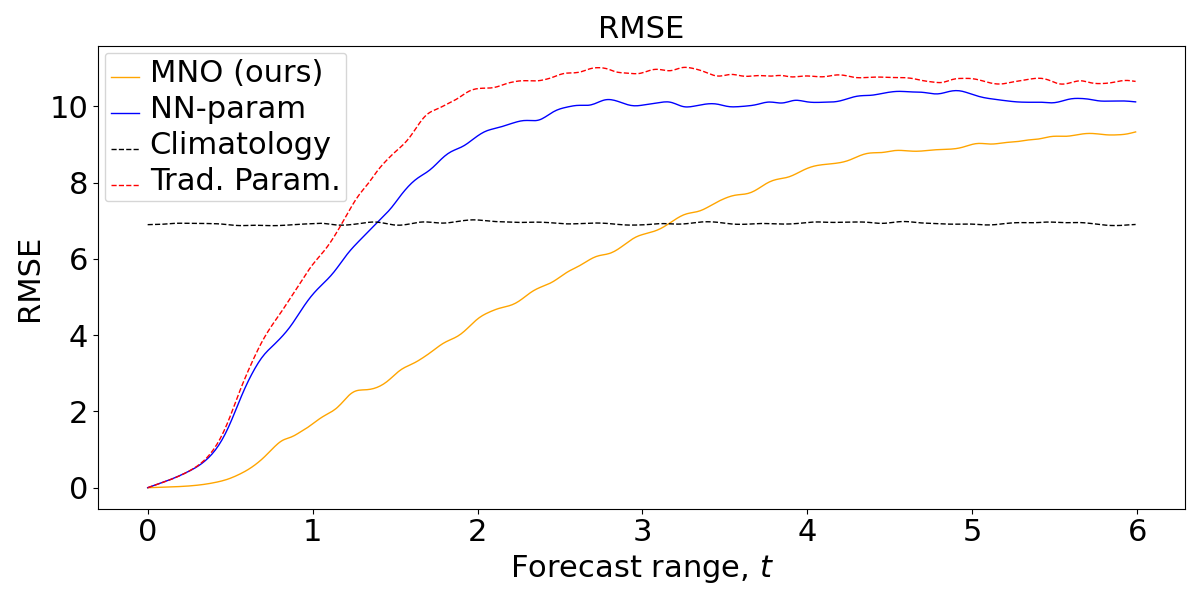}}
\caption[Stability]{\textbf{MNO is stable.} MNO can propagate a sample state, $X_{k=0}(t)$, over a long time horizon without diverging to infinity (left). The right plot shows that the RMSE of MNO plateaus for long-term forecasts, further confirming stability. Further, MNO (yellow) maintains accuracy longer than ML-based parametrizations (blue) and a climatology (black).}
\label{fig:stability}
\end{figure}

\Cref{fig:stability} shows that predicting large-scale dynamics with MNO is stable. We first plot a randomly selected sample of the first large-scale state, $X_{k=0}(t)$  (left-black), to illustrate that the prediction is bounded. The MNO prediction (left-yellow) follows the ground-truth up to an approximate horizon of, $t=1.8$ or $9$ days, then diverges from the ground-truth solution, but stays within the bounds of the ground-truth prediction and does not diverge to infinity. The RMSE over time in~\Cref{fig:stability} shows that MNO (yellow) is approximately more accurate than current ML-based (blue) and traditional (red) parametrizations for $\approx 100\%$-longer time, measuring the time to intersect with climatology. Despite the difficulty in predicting chaotic dynamics, the RMSE of MNO reaches a plateau, which is slightly above the optimal plateau given by the climatology (black).

The RMSE over time is calculated as: 
\begin{equation}
\begin{aligned}
\text{RMSE}(t) &= \frac{1}{K} \sum_{k=0}^K \sqrt(\frac{1}{N}\sum_{i=0}^N(\hat X_{k,i}(t) - X_{k,i}(t))^2).
\end{aligned}
\label{eq:rmse}
\end{equation}

\section{Limitations and Future Work}
We demonstrated the accuracy, speed, and stability of MNO on the chaotic multiscale Lorenz96 equation. 
Future work, can extend MNO towards higher-dimensional or time-irregular systems and further integrate symmetries or constraints:

The results show promise to extend MNO to higher-dimensional, chaotic, multiscale, multiphysics problems and improve parametrizations in anisotropic turbulence predictions~\citep{sagaut06les}, Rayleigh-B\'enard Convection (see Appenix A.1.) 
or clouds of global atmospheric models~\citep{wang22nonlocal,Palmer_2019}. Lightweight climate surrogate models could dramatically improve uncertainties~\citep{lutjens2021spectral} or decision-exploration~\citep{rooneyvarge20enroads} in climate.

MNO is grid-independent in space but not in time which could be alleviated via integrations with Neural ODEs~\citep{Chen_2018}. MNO is a myopic model which might suffice for chaotic dynamics~\citep{li21markov}, but could be combined with LSTMs~\citep{mohan19turbulencelstm} or reservoir computing~\citep{pathak18reservoircomputing} to contain a memory. Further, we leveraged global Fourier decompositions to exploit grid-independent periodic spatial correlations, but future work could also capture local discontinuities, e.g., along coastlines~\citep{jiang21dte} with multiwavelets~\citep{gupta2021multiwavelet}, or incorporate non-periodic boundaries via Chebyshev polynomials. 

Lastly, MNO can be combined with Geometric deep learning, PINNs, or hard constraint models. This avenue of research is particularly exciting with MNO as there exist many known symmetries for the paramtrization term~\citep{prakash21les}. 



\section{Conclusion}
We proposed a hybrid physics-ML surrogate of multiscale PDEs that is quasilinear, accurate, and stable. The surrogate limits learning to the influence of fine- onto large-scale dynamics and is the first to use neural operators for a grid-independent, non-local corrective term of large-scale simulations. We demonstrated that multiscale neural operator (MNO) is faster than direct numerical simulation ($O(N\log N)$ vs. $O(N^2$) and more accurate ($\approx100\%$ longer prediction horizon) than state-of-the-art parametrizations on the chaotic, multiscale equations multiscale Lorenz96. With the dramatic reduction in runtime, MNO could enable rapid parameter exploration and robust uncertainty quantification in complex climate models. 


\section{Ethical and Societal Implications of the proposed work}\label{sec:ethics}
Climate change is the defining challenge of our time. Environmental disasters will become more frequent: from storms, floods, wildfires and heat waves to biodiversity loss and air pollution~\citep{IPCC_2018}. The impacts of climate change will not only be severe, but also unjustly distributed: island states, minority populations, and the Global South are already facing the most severe consequences of climate change, while the Global North is responsible for the most emissions since the industrial revolution~\citep{Althor_2016}.  
Decision-makers require better tools to understand and plan for changes in climate and limit the economic, human, and environmental impact~\citep{Palmer_2019}. We propose a faster differential equation solver to improve the underlying climate models. Because fast differential equations can be leveraged in ethically questionable fields, such as missile development, we are applying our methods to climate modeling to demonstrate our work towards positive impact.

\bibliographystyle{plainnat}
\bibliography{references}

\begin{thebibliography}{145}
\providecommand{\natexlab}[1]{#1}
\providecommand{\url}[1]{\texttt{#1}}
\expandafter\ifx\csname urlstyle\endcsname\relax
  \providecommand{\doi}[1]{doi: #1}\else
  \providecommand{\doi}{doi: \begingroup \urlstyle{rm}\Url}\fi

\bibitem[Althor et~al.(2016)Althor, Watson, and Fuller]{Althor_2016}
Glenn Althor, James E.~M. Watson, and Richard~A. Fuller.
\newblock Global mismatch between greenhouse gas emissions and the burden of
  climate change.
\newblock \emph{Scientific Reports}, 6, 2016.

\bibitem[Anandkumar et~al.(2020)Anandkumar, Azizzadenesheli, Bhattacharya,
  Kovachki, Li, Liu, and Stuart]{Anandkumar_2020}
Anima Anandkumar, Kamyar Azizzadenesheli, Kaushik Bhattacharya, Nikola
  Kovachki, Zongyi Li, Burigede Liu, and Andrew Stuart.
\newblock Neural operator: Graph kernel network for partial differential
  equations.
\newblock In \emph{ICLR 2020 Workshop on Integration of Deep Neural Models and
  Differential Equations}, 2020.

\bibitem[Batzner et~al.(2022)Batzner, Musaelian, Sun, Geiger, Mailoa,
  Kornbluth, Molinari, Smidt, and Kozinsky]{batzner22equivariantmolecule}
Simon Batzner, Albert Musaelian, Lixin Sun, Mario Geiger, Jonathan~P. Mailoa,
  Mordechai Kornbluth, Nicola Molinari, Tess~E. Smidt, and Boris Kozinsky.
\newblock E(3)-equivariant graph neural networks for data-efficient and
  accurate interatomic potentials.
\newblock \emph{Nature Communications}, 13, 2022.

\bibitem[Behler(2011)]{behler11chem}
Jörg Behler.
\newblock Neural network potential-energy surfaces in chemistry: a tool for
  large-scale simulations.
\newblock \emph{Phys. Chem. Chem. Phys.}, 13, 2011.

\bibitem[Behler~J(2007)]{behler07chemparam}
Parrinello~M. Behler~J.
\newblock Generalized neural-network representation of high-dimensional
  potential-energy surfaces.
\newblock \emph{Phys Rev Lett.}, 14, 2007.

\bibitem[Bennett and Nijssen(2020)]{bennett20hydroparams}
Andrew Bennett and Bart Nijssen.
\newblock Deep learned process parameterizations provide better representations
  of turbulent heat fluxes in hydrologic models.
\newblock \emph{Earth and Space Science Open Archive}, page~20, 2020.

\bibitem[{Beucler} et~al.(2019){Beucler}, {Rasp}, {Pritchard}, and
  {Gentine}]{beucler19energycons}
Tom {Beucler}, Stephan {Rasp}, Michael {Pritchard}, and Pierre {Gentine}.
\newblock {Achieving Conservation of Energy in Neural Network Emulators for
  Climate Modeling}.
\newblock jun 2019.

\bibitem[Beucler et~al.(2021{\natexlab{a}})Beucler, Pritchard, Rasp, Ott,
  Baldi, and Gentine]{beucler21constraints}
Tom Beucler, Michael Pritchard, Stephan Rasp, Jordan Ott, Pierre Baldi, and
  Pierre Gentine.
\newblock Enforcing analytic constraints in neural networks emulating physical
  systems.
\newblock \emph{Phys. Rev. Lett.}, 126:\penalty0 098302, Mar
  2021{\natexlab{a}}.

\bibitem[Beucler et~al.(2021{\natexlab{b}})Beucler, Pritchard, Yuval, Gupta,
  Peng, Rasp, Ahmed, O'Gorman, Neelin, Lutsko, and Gentine]{beucler21ciml}
Tom Beucler, Michael~S. Pritchard, Janni Yuval, Ankitesh Gupta, Liran Peng,
  Stephan Rasp, Fiaz Ahmed, Paul~A. O'Gorman, J.~David Neelin, Nicholas~J.
  Lutsko, and Pierre Gentine.
\newblock Climate-invariant machine learning.
\newblock \emph{CoRR}, 2021{\natexlab{b}}.

\bibitem[Bhattacharya et~al.(2020)Bhattacharya, Hosseini, Kovachki, and
  Stuart]{Bhattacharya_2020}
Kaushik Bhattacharya, Bamdad Hosseini, Nikola~B. Kovachki, and Andrew~M.
  Stuart.
\newblock Model reduction and neural networks for parametric pdes, 2020.

\bibitem[Bieker et~al.(2020)Bieker, Peitz, Brunton, Kutz, and
  Dellnitz]{bieker20control}
Katharina Bieker, Sebastian Peitz, Steven~L. Brunton, J.~Nathan Kutz, and
  Michael Dellnitz.
\newblock Deep model predictive flow control with limited sensor data and
  online learning, 2020.

\bibitem[Blakseth et~al.(2022)Blakseth, Rasheed, Kvamsdal, and
  San]{blakseth22resfcnns}
Sindre~Stenen Blakseth, Adil Rasheed, Trond Kvamsdal, and Omer San.
\newblock Deep neural network enabled corrective source term approach to hybrid
  analysis and modeling.
\newblock \emph{Neural Networks}, 146:\penalty0 181--199, 2022.

\bibitem[Bode et~al.(2021)Bode, Gauding, Lian, Denker, Davidovic, Kleinheinz,
  Jitsev, and Pitsch]{bode21pinnsuperresolution}
Mathis Bode, Michael Gauding, Zeyu Lian, Dominik Denker, Marco Davidovic,
  Konstantin Kleinheinz, Jenia Jitsev, and Heinz Pitsch.
\newblock Using physics-informed enhanced super-resolution generative
  adversarial networks for subfilter modeling in turbulent reactive flows.
\newblock \emph{Proceedings of the Combustion Institute}, 38\penalty0
  (2):\penalty0 2617--2625, 2021.

\bibitem[Bonavita and Laloyaux(2020)]{bonavita20mlassim}
Massimo Bonavita and Patrick Laloyaux.
\newblock Machine learning for model error inference and correction.
\newblock \emph{Journal of Advances in Modeling Earth Systems}, 12\penalty0
  (12), 2020.

\bibitem[Boyd(2013)]{boyd13chebyshev}
J.P. Boyd.
\newblock \emph{Chebyshev and Fourier Spectral Methods: Second Revised
  Edition}.
\newblock Dover Books on Mathematics. Dover Publications, 2013.

\bibitem[Brandstetter et~al.(2022)Brandstetter, Worrall, and
  Welling]{brandstetter2022gnnpdes}
Johannes Brandstetter, Daniel~E. Worrall, and Max Welling.
\newblock Message passing neural {PDE} solvers.
\newblock In \emph{International Conference on Learning Representations
  (ICLR)}, 2022.

\bibitem[Brenowitz and Bretherton(2018)]{brenowitz18param}
N.~D. Brenowitz and C.~S. Bretherton.
\newblock Prognostic validation of a neural network unified physics
  parameterization.
\newblock \emph{Geophysical Research Letters}, 45\penalty0 (12):\penalty0
  6289--6298, 2018.

\bibitem[Brenowitz et~al.(2020)Brenowitz, Beucler, Pritchard, and
  Bretherton]{brenowitz20interpretparams}
Noah~D. Brenowitz, Tom Beucler, Michael Pritchard, and Christopher~S.
  Bretherton.
\newblock Interpreting and stabilizing machine-learning parametrizations of
  convection.
\newblock \emph{Journal of the Atmospheric Sciences}, 77\penalty0
  (12):\penalty0 4357 -- 4375, 2020.

\bibitem[Bretherton et~al.(2022)Bretherton, Henn, Kwa, Brenowitz, Watt-Meyer,
  McGibbon, Perkins, Clark, and Harris]{bretherton22coarse}
Christopher~S. Bretherton, Brian Henn, Anna Kwa, Noah~D. Brenowitz, Oliver
  Watt-Meyer, Jeremy McGibbon, W.~Andre Perkins, Spencer~K. Clark, and Lucas
  Harris.
\newblock Correcting coarse-grid weather and climate models by machine learning
  from global storm-resolving simulations.
\newblock \emph{Journal of Advances in Modeling Earth Systems}, 14\penalty0
  (2), 2022.

\bibitem[Briggs et~al.(2000)Briggs, Henson, and McCormick]{briggs00multigrid}
William~L. Briggs, Van~Emden Henson, and Steve~F. McCormick.
\newblock \emph{A Multigrid Tutorial (2nd Ed.)}.
\newblock Society for Industrial and Applied Mathematics, USA, 2000.
\newblock ISBN 0898714621.

\bibitem[Bronstein et~al.(2021)Bronstein, Bruna, Cohen, and
  Velickovic]{bronstein21geometric}
Michael~M. Bronstein, Joan Bruna, Taco Cohen, and Petar Velickovic.
\newblock Geometric deep learning: Grids, groups, graphs, geodesics, and
  gauges.
\newblock \emph{CoRR}, 2021.

\bibitem[Brunton et~al.(2016)Brunton, Proctor, and Kutz]{brunton16sindy}
Steven~L. Brunton, Joshua~L. Proctor, and J.~Nathan Kutz.
\newblock Discovering governing equations from data by sparse identification of
  nonlinear dynamical systems.
\newblock \emph{Proceedings of the National Academy of Sciences}, 113\penalty0
  (15):\penalty0 3932--3937, 2016.

\bibitem[{Burns} et~al.(2020){Burns}, {Vasil}, {Oishi}, {Lecoanet}, and
  {Brown}]{burns20dedalus}
Keaton~J. {Burns}, Geoffrey~M. {Vasil}, Jeffrey~S. {Oishi}, Daniel {Lecoanet},
  and Benjamin~P. {Brown}.
\newblock {Dedalus: A flexible framework for numerical simulations with
  spectral methods}.
\newblock \emph{Physical Review Research}, 2\penalty0 (2), April 2020.

\bibitem[Burns et~al.(2022)Burns, Eyink, Meneveau, Szalay, Zaki, Vishniac,
  Gupta, Wang, Hao, Wu, and Lemson]{jh22turbulencedata}
Randal Burns, Gregory Eyink, Charles Meneveau, Alex Szalay, Tamer Zaki, Ethan
  Vishniac, Akshat Gupta, Mengze Wang, Yue Hao, Zhao Wu, and Gerard Lemson.
\newblock Johns hopkins turbulence database, 2022.
\newblock last accessed May, 2022.

\bibitem[Cachay et~al.(2021{\natexlab{a}})Cachay, Erickson, Bucker, Pokropek,
  Potosnak, Bire, Osei, and L\"utjens]{cachay21graphino}
Salva~R\"uhling Cachay, Emma Erickson, Arthur Fender~C. Bucker, Ernest
  Pokropek, Willa Potosnak, Suyash Bire, Salomey Osei, and Bj\"orn L\"utjens.
\newblock The world as a graph: Improving el ni\~no forecasts with graph neural
  networks, 2021{\natexlab{a}}.

\bibitem[Cachay et~al.(2021{\natexlab{b}})Cachay, Ramesh, Cole, Barker, and
  Rolnick]{cachay21climart}
Salva~R{\"u}hling Cachay, Venkatesh Ramesh, Jason N.~S. Cole, Howard Barker,
  and David Rolnick.
\newblock Clim{ART}: A benchmark dataset for emulating atmospheric radiative
  transfer in weather and climate models.
\newblock In \emph{Thirty-fifth Conference on Neural Information Processing
  Systems Datasets and Benchmarks Track (Round 2)}, 2021{\natexlab{b}}.

\bibitem[Campin et~al.(2011)Campin, Hill, Jones, and Marshall]{Campin_2011}
J.~Campin, C.~Hill, H.~Jones, and J.~Marshall.
\newblock Super-parameterization in ocean modeling: Application to deep
  convection.
\newblock \emph{Ocean Modelling}, 36:\penalty0 90--101, 2011.

\bibitem[Carleo et~al.(2019)Carleo, Cirac, Cranmer, Daudet, Schuld, Tishby,
  Vogt-Maranto, and Zdeborov\'a]{carleo19mlps}
Giuseppe Carleo, Ignacio Cirac, Kyle Cranmer, Laurent Daudet, Maria Schuld,
  Naftali Tishby, Leslie Vogt-Maranto, and Lenka Zdeborov\'a.
\newblock Machine learning and the physical sciences.
\newblock \emph{Rev. Mod. Phys.}, 91, Dec 2019.

\bibitem[Chakraborty et~al.(2021)Chakraborty, Adhikari, and
  Ganguli]{chakraborty21gps}
S.~Chakraborty, S.~Adhikari, and R.~Ganguli.
\newblock The role of surrogate models in the development of digital twins of
  dynamic systems.
\newblock \emph{Applied Mathematical Modelling}, 90:\penalty0 662--681, 2021.

\bibitem[Chan and Elsheikh(2019)]{chan19geological}
Shing Chan and Ahmed~H. Elsheikh.
\newblock Parametric generation of conditional geological realizations using
  generative neural networks, 2019.

\bibitem[Chatterjee(2000)]{chatterjee00pod}
Anindya Chatterjee.
\newblock An introduction to the proper orthogonal decomposition.
\newblock \emph{Current Science}, 78\penalty0 (7):\penalty0 808--817, 2000.

\bibitem[Chattopadhyay et~al.(2020{\natexlab{a}})Chattopadhyay, Mustafa,
  Hassanzadeh, and Kashinath]{chattopadhyay20transformers}
Ashesh Chattopadhyay, Mustafa Mustafa, Pedram Hassanzadeh, and Karthik
  Kashinath.
\newblock Deep spatial transformers for autoregressive data-driven forecasting
  of geophysical turbulence.
\newblock In \emph{Proceedings of the 10th International Conference on Climate
  Informatics}, CI2020, page 106–112, New York, NY, USA, 2020{\natexlab{a}}.
  Association for Computing Machinery.

\bibitem[Chattopadhyay et~al.(2020{\natexlab{b}})Chattopadhyay, Subel, and
  Hassanzadeh]{chattopadhyay20lorenz}
Ashesh Chattopadhyay, Adam Subel, and Pedram Hassanzadeh.
\newblock Data-driven super-parameterization using deep learning:
  Experimentation with multiscale lorenz 96 systems and transfer learning.
\newblock \emph{Journal of Advances in Modeling Earth Systems}, 12\penalty0
  (11), 2020{\natexlab{b}}.

\bibitem[Chen et~al.(2018)Chen, Rubanova, Bettencourt, and Duvenaud]{Chen_2018}
Tian~Qi Chen, Yulia Rubanova, Jesse Bettencourt, and David~K Duvenaud.
\newblock Neural ordinary differential equations.
\newblock In \emph{Advances in Neural Information Processing Systems 31}, pages
  6571--6583. Curran Associates, Inc., 2018.

\bibitem[Chen and Chen(1995)]{chen95universaloperator}
Tianping Chen and Hong Chen.
\newblock Universal approximation to nonlinear operators by neural networks
  with arbitrary activation functions and its application to dynamical systems.
\newblock \emph{IEEE Transactions on Neural Networks}, 6\penalty0 (4):\penalty0
  911--917, 1995.

\bibitem[Chen and Ahmed()]{chen20padgan}
Wei Chen and Faez Ahmed.
\newblock Padgan: Learning to generate high-quality novel designs.
\newblock \emph{Journal of Mechanical Design}, 143\penalty0 (3).

\bibitem[Cohen and Welling(2017)]{cohen17steerablecnns}
Taco~S. Cohen and Max Welling.
\newblock Steerable cnns.
\newblock In \emph{5th International Conference on Learning Representations,
  {ICLR} 2017, Toulon, France}, 2017.

\bibitem[Costa~Nogueira et~al.(2020)Costa~Nogueira, de~Sousa~Almeida, Auger,
  and Watson]{Costa_2020}
Alberto Costa~Nogueira, Jo{\~a}o~Lucas de~Sousa~Almeida, Guillaume Auger, and
  Campbell~D. Watson.
\newblock Reduced order modeling of dynamical systems using artificial neural
  networks applied to water circulation.
\newblock In Heike Jagode, Hartwig Anzt, Guido Juckeland, and Hatem Ltaief,
  editors, \emph{High Performance Computing}, pages 116--136, Cham, 2020.
  Springer International Publishing.

\bibitem[Donti et~al.(2021)Donti, Rolnick, and Kolter]{donti2021dc3}
Priya~L. Donti, David Rolnick, and J~Zico Kolter.
\newblock {DC}3: A learning method for optimization with hard constraints.
\newblock In \emph{International Conference on Learning Representations
  (ICLR)}, 2021.

\bibitem[Dueben and Bauer(2018)]{dueben18nonlocalparams}
P.~D. Dueben and P.~Bauer.
\newblock Challenges and design choices for global weather and climate models
  based on machine learning.
\newblock \emph{Geoscientific Model Development}, 11\penalty0 (10):\penalty0
  3999--4009, 2018.

\bibitem[Duraisamy et~al.(2019)Duraisamy, Iaccarino, and
  Xiao]{duraisamy19turbreview}
Karthik Duraisamy, Gianluca Iaccarino, and Heng Xiao.
\newblock Turbulence modeling in the age of data.
\newblock \emph{Annual Review of Fluid Mechanics}, 51\penalty0 (1):\penalty0
  357--377, 2019.

\bibitem[Dutt and Rokhlin(1993)]{dutt93nudft}
A.~Dutt and V.~Rokhlin.
\newblock Fast fourier transforms for nonequispaced data.
\newblock \emph{SIAM Journal on Scientific Computing}, 14\penalty0
  (6):\penalty0 1368--1393, 1993.

\bibitem[Farlow(1993)]{farlow93pdes}
S.J. Farlow.
\newblock \emph{Partial Differential Equations for Scientists and Engineers}.
\newblock Dover books on advanced mathematics. Dover Publications, 1993.

\bibitem[Fuchs et~al.(2020)Fuchs, Worrall, Fischer, and Welling]{fuchs20se3}
Fabian Fuchs, Daniel Worrall, Volker Fischer, and Max Welling.
\newblock Se(3)-transformers: 3d roto-translation equivariant attention
  networks.
\newblock In H.~Larochelle, M.~Ranzato, R.~Hadsell, M.F. Balcan, and H.~Lin,
  editors, \emph{Advances in Neural Information Processing Systems}, volume~33.
  Curran Associates, Inc., 2020.

\bibitem[Fuhrer et~al.(2018)Fuhrer, Chadha, Hoefler, Kwasniewski, Lapillonne,
  Leutwyler, L\"uthi, Osuna, Sch\"ar, Schulthess, and Vogt]{Fuhrer_2018}
O.~Fuhrer, T.~Chadha, T.~Hoefler, G.~Kwasniewski, X.~Lapillonne, D.~Leutwyler,
  D.~L\"uthi, C.~Osuna, C.~Sch\"ar, T.~C. Schulthess, and H.~Vogt.
\newblock Near-global climate simulation at 1 km resolution: establishing a
  performance baseline on 4888 gpus with cosmo 5.0.
\newblock \emph{Geosci. Model Dev.}, 11:\penalty0 1665 -- 1681, 2018.

\bibitem[Fukunaga and Koontz(1970)]{fukunaga70kle}
K.~Fukunaga and W.L.G. Koontz.
\newblock Application of the karhunen-loève expansion to feature selection and
  ordering.
\newblock \emph{IEEE Transactions on Computers}, C-19\penalty0 (4):\penalty0
  311--318, 1970.

\bibitem[Ganguly et~al.(2014)Ganguly, Kodra, Agrawal, Banerjee, Boriah,
  Chatterjee, Chatterjee, Choudhary, Das, Faghmous, Ganguli, Ghosh, Hayhoe,
  Hays, Hendrix, Fu, Kawale, Kumar, Kumar, Liao, Liess, Mawalagedara, Mithal,
  Oglesby, Salvi, Snyder, Steinhaeuser, Wang, and
  Wuebbles]{ganguly14weathermlreview}
A.~R. Ganguly, E.~A. Kodra, A.~Agrawal, A.~Banerjee, S.~Boriah, Sn. Chatterjee,
  So. Chatterjee, A.~Choudhary, D.~Das, J.~Faghmous, P.~Ganguli, S.~Ghosh,
  K.~Hayhoe, C.~Hays, W.~Hendrix, Q.~Fu, J.~Kawale, D.~Kumar, V.~Kumar,
  W.~Liao, S.~Liess, R.~Mawalagedara, V.~Mithal, R.~Oglesby, K.~Salvi, P.~K.
  Snyder, K.~Steinhaeuser, D.~Wang, and D.~Wuebbles.
\newblock Toward enhanced understanding and projections of climate extremes
  using physics-guided data mining techniques.
\newblock \emph{Nonlinear Processes in Geophysics}, 21\penalty0 (4):\penalty0
  777--795, 2014.

\bibitem[Gentine et~al.(2018)Gentine, Pritchard, Rasp, Reinaudi, and
  Yacalis]{gentine18params}
P.~Gentine, M.~Pritchard, S.~Rasp, G.~Reinaudi, and G.~Yacalis.
\newblock Could machine learning break the convection parameterization
  deadlock?
\newblock \emph{Geophysical Research Letters}, 45\penalty0 (11):\penalty0
  5742--5751, 2018.

\bibitem[Goodfellow et~al.(2014)Goodfellow, Pouget-Abadie, Mirza, Xu,
  Warde-Farley, Ozair, Courville, and Bengio]{goodfellow14gans}
Ian Goodfellow, Jean Pouget-Abadie, Mehdi Mirza, Bing Xu, David Warde-Farley,
  Sherjil Ozair, Aaron Courville, and Yoshua Bengio.
\newblock Generative adversarial nets.
\newblock In \emph{Advances in Neural Information Processing Systems},
  volume~27. Curran Associates, Inc., 2014.

\bibitem[Greydanus et~al.(2019)Greydanus, Dzamba, and
  Yosinski]{greydanus19hamiltonian}
Samuel Greydanus, Misko Dzamba, and Jason Yosinski.
\newblock Hamiltonian neural networks.
\newblock In H.~Wallach, H.~Larochelle, A.~Beygelzimer, F.~d~Alch\'{e}-Buc,
  E.~Fox, and R.~Garnett, editors, \emph{Advances in Neural Information
  Processing Systems 32}, pages 15379--15389. Curran Associates, Inc., 2019.

\bibitem[Groenke et~al.(2020)Groenke, Madaus, and
  Monteleoni]{groenke20climalign}
Brian Groenke, Luke Madaus, and Claire Monteleoni.
\newblock Climalign: Unsupervised statistical downscaling of climate variables
  via normalizing flows.
\newblock In \emph{Proceedings of the 10th International Conference on Climate
  Informatics}, CI2020, page 60–66, New York, NY, USA, 2020. Association for
  Computing Machinery.

\bibitem[Gupta et~al.(2021)Gupta, Xiao, and Bogdan]{gupta2021multiwavelet}
Gaurav Gupta, Xiongye Xiao, and Paul Bogdan.
\newblock Multiwavelet-based operator learning for differential equations.
\newblock In A.~Beygelzimer, Y.~Dauphin, P.~Liang, and J.~Wortman Vaughan,
  editors, \emph{Advances in Neural Information Processing Systems (NeurIPS)},
  2021.

\bibitem[Hamilton et~al.(2017)Hamilton, Lloyd, and Flores]{hamilton17inverse}
Franz Hamilton, Alun~L. Lloyd, and Kevin~B. Flores.
\newblock Hybrid modeling and prediction of dynamical systems.
\newblock \emph{PLOS Computational Biology}, 13\penalty0 (7):\penalty0 1--20,
  07 2017.

\bibitem[Hansen et~al.(2013)Hansen, Montavon, Biegler, Fazli, Rupp, Scheffler,
  von Lilienfeld, Tkatchenko, and Müller]{hansen13chemparam}
Katja Hansen, Grégoire Montavon, Franziska Biegler, Siamac Fazli, Matthias
  Rupp, Matthias Scheffler, O.~Anatole von Lilienfeld, Alexandre Tkatchenko,
  and Klaus-Robert Müller.
\newblock Assessment and validation of machine learning methods for predicting
  molecular atomization energies.
\newblock \emph{Journal of Chemical Theory and Computation}, 9\penalty0
  (8):\penalty0 3404--3419, 2013.

\bibitem[Hasani et~al.(2021)Hasani, Lechner, Amini, Rus, and
  Grosu]{hasani21ltc}
Ramin Hasani, Mathias Lechner, Alexander Amini, Daniela Rus, and Radu Grosu.
\newblock Liquid time-constant networks.
\newblock \emph{Proceedings of the AAAI Conference on Artificial Intelligence},
  35\penalty0 (9):\penalty0 7657--7666, May 2021.

\bibitem[Hersbach et~al.(2020)Hersbach, Bell, Berrisford, Hirahara, Horányi,
  Muñoz-Sabater, Nicolas, Peubey, Radu, Schepers, Simmons, Soci, Abdalla,
  Abellan, Balsamo, Bechtold, Biavati, Bidlot, Bonavita, De~Chiara, Dahlgren,
  Dee, Diamantakis, Dragani, Flemming, Forbes, Fuentes, Geer, Haimberger,
  Healy, Hogan, Hólm, Janisková, Keeley, Laloyaux, Lopez, Lupu, Radnoti,
  de~Rosnay, Rozum, Vamborg, Villaume, and Thépaut]{herbach20era5}
Hans Hersbach, Bill Bell, Paul Berrisford, Shoji Hirahara, András Horányi,
  Joaquín Muñoz-Sabater, Julien Nicolas, Carole Peubey, Raluca Radu, Dinand
  Schepers, Adrian Simmons, Cornel Soci, Saleh Abdalla, Xavier Abellan,
  Gianpaolo Balsamo, Peter Bechtold, Gionata Biavati, Jean Bidlot, Massimo
  Bonavita, Giovanna De~Chiara, Per Dahlgren, Dick Dee, Michail Diamantakis,
  Rossana Dragani, Johannes Flemming, Richard Forbes, Manuel Fuentes, Alan
  Geer, Leo Haimberger, Sean Healy, Robin~J. Hogan, Elías Hólm, Marta
  Janisková, Sarah Keeley, Patrick Laloyaux, Philippe Lopez, Cristina Lupu,
  Gabor Radnoti, Patricia de~Rosnay, Iryna Rozum, Freja Vamborg, Sebastien
  Villaume, and Jean-Noël Thépaut.
\newblock The era5 global reanalysis.
\newblock \emph{Quarterly Journal of the Royal Meteorological Society},
  146\penalty0 (730):\penalty0 1999--2049, 2020.

\bibitem[IPCC(2018)]{IPCC_2018}
IPCC.
\newblock Global warming of 1.5c. an ipcc special report on the impacts of
  global warming of 1.5c above pre-industrial levels and related global
  greenhouse gas emission pathways, in the context of strengthening the global
  response to the threat of climate change, sustainable development, and
  efforts to eradicate poverty, 2018.

\bibitem[Jia et~al.(2019)Jia, Willard, Karpatne, Read, Zwart, Steinbach, and
  Kumar]{jia19pgrnn}
Xiaowei Jia, Jared Willard, Anuj Karpatne, Jordan Read, Jacob Zwart, {Michael
  S} Steinbach, and Vipin Kumar.
\newblock Physics guided rnns for modeling dynamical systems: A case study in
  simulating lake temperature profiles.
\newblock In \emph{SIAM International Conference on Data Mining, SDM 2019},
  SIAM International Conference on Data Mining, SDM 2019, pages 558--566.
  Society for Industrial and Applied Mathematics Publications, 2019.

\bibitem[Jia et~al.(2021)Jia, Willard, Karpatne, Read, Zwart, Steinbach, and
  Kumar]{jia21pgml}
Xiaowei Jia, Jared Willard, Anuj Karpatne, Jordan~S. Read, Jacob~A. Zwart,
  Michael Steinbach, and Vipin Kumar.
\newblock Physics-guided machine learning for scientific discovery: An
  application in simulating lake temperature profiles.
\newblock \emph{ACM/IMS Trans. Data Sci.}, 2\penalty0 (3), 2021.

\bibitem[{Jiang} et~al.(2021){Jiang}, {Meinert}, {Jord{\~a}o}, {Weisser},
  {Holgate}, {Lavin}, {L{\"u}tjens}, {Newman}, {Wainwright}, {Walker}, and
  {Barnard}]{jiang21dte}
Peishi {Jiang}, Nis {Meinert}, Helga {Jord{\~a}o}, Constantin {Weisser}, Simon
  {Holgate}, Alexander {Lavin}, Bj{\"o}rn {L{\"u}tjens}, Dava {Newman}, Haruko
  {Wainwright}, Catherine {Walker}, and Patrick {Barnard}.
\newblock {Digital Twin Earth -- Coasts: Developing a fast and physics-informed
  surrogate model for coastal floods via neural operators}.
\newblock \emph{2021 NeurIPS Workshop on Machine Learning for the Physical
  Sciences (ML4PS)}, 2021.

\bibitem[Jin et~al.(2020)Jin, Zhang, Zhu, Tang, and
  Karniadakis]{pengzhan20sympnet}
Pengzhan Jin, Zhen Zhang, Aiqing Zhu, Yifa Tang, and George~Em Karniadakis.
\newblock Sympnets: Intrinsic structure-preserving symplectic networks for
  identifying hamiltonian systems.
\newblock \emph{Neural Networks}, 132, 12 2020.

\bibitem[Kani and Elsheikh(2017)]{kani17drrnn}
J.~Nagoor Kani and Ahmed~H. Elsheikh.
\newblock {DR-RNN:} {A} deep residual recurrent neural network for model
  reduction.
\newblock \emph{CoRR}, abs/1709.00939, 2017.

\bibitem[Karniadakis et~al.(2021)Karniadakis, Kevrekidis, Lu, Perdikaris, Wang,
  and Yang]{Karniadakis_2021}
George~Em Karniadakis, Ioannis~G. Kevrekidis, Lu~Lu, Paris Perdikaris, Sifan
  Wang, and Liu Yang.
\newblock Physics-informed machine learning.
\newblock \emph{Nature Reviews Physics}, 3:\penalty0 422--440, June 2021.

\bibitem[{Karpatne} et~al.(2019){Karpatne}, {Ebert-Uphoff}, {Ravela}, {Babaie},
  and {Kumar}]{Karpatne_2019}
A.~{Karpatne}, I.~{Ebert-Uphoff}, S.~{Ravela}, H.~A. {Babaie}, and V.~{Kumar}.
\newblock Machine learning for the geosciences: Challenges and opportunities.
\newblock \emph{IEEE Transactions on Knowledge and Data Engineering},
  31\penalty0 (8):\penalty0 1544--1554, Aug 2019.

\bibitem[Karpatne et~al.(2017)Karpatne, Atluri, Faghmous, Steinbach, Banerjee,
  Ganguly, Shekhar, Samatova, and Kumar]{karpatne17review}
Anuj Karpatne, Gowtham Atluri, James~H. Faghmous, Michael Steinbach, Arindam
  Banerjee, Auroop Ganguly, Shashi Shekhar, Nagiza Samatova, and Vipin Kumar.
\newblock Theory-guided data science: A new paradigm for scientific discovery
  from data.
\newblock \emph{IEEE Transactions on Knowledge and Data Engineering},
  29\penalty0 (10):\penalty0 2318--2331, 2017.

\bibitem[{Karpatne} et~al.(2017){Karpatne}, {Watkins}, {Read}, and
  {Kumar}]{karpatne17pgnn}
Anuj {Karpatne}, William {Watkins}, Jordan {Read}, and Vipin {Kumar}.
\newblock {Physics-guided Neural Networks (PGNN): An Application in Lake
  Temperature Modeling}.
\newblock \emph{arXiv e-prints}, page arXiv:1710.11431, October 2017.

\bibitem[Kashinath et~al.(2021)Kashinath, Mustafa, Albert, Wu, Jiang,
  Esmaeilzadeh, Azizzadenesheli, Wang, Chattopadhyay, Singh, Manepalli,
  Chirila, Yu, Walters, White, Xiao, Tchelepi, Marcus, Anandkumar, Hassanzadeh,
  and Prabhat]{kashinath21review}
K.~Kashinath, M.~Mustafa, A.~Albert, J-L. Wu, C.~Jiang, S.~Esmaeilzadeh,
  K.~Azizzadenesheli, R.~Wang, A.~Chattopadhyay, A.~Singh, A.~Manepalli,
  D.~Chirila, R.~Yu, R.~Walters, B.~White, H.~Xiao, H.~A. Tchelepi, P.~Marcus,
  A.~Anandkumar, P.~Hassanzadeh, and null Prabhat.
\newblock Physics-informed machine learning: case studies for weather and
  climate modelling.
\newblock \emph{Philosophical Transactions of the Royal Society A:
  Mathematical, Physical and Engineering Sciences}, 379\penalty0 (2194), 2021.

\bibitem[Khairoutdinov et~al.(2005)Khairoutdinov, Randall, and
  DeMott]{Khairoutdinov_2005}
Marat Khairoutdinov, David Randall, and Charlotte DeMott.
\newblock {Simulations of the atmospheric general circulation using a
  cloud-resolving model as a superparameterization of physical processes}.
\newblock \emph{Journal of the Atmospheric Sciences}, 62\penalty0 (7
  I):\penalty0 2136--2154, jul 2005.

\bibitem[Khoo and Ying(2019)]{khoo19switchnet}
Yuehaw Khoo and Lexing Ying.
\newblock Switchnet: A neural network model for forward and inverse scattering
  problems.
\newblock \emph{SIAM Journal on Scientific Computing}, 41\penalty0 (5), 2019.

\bibitem[Kingma and Ba(2015)]{Kingma_2015}
Diederik~P. Kingma and Jimmy Ba.
\newblock Adam: {A} method for stochastic optimization.
\newblock In Yoshua Bengio and Yann LeCun, editors, \emph{3rd International
  Conference on Learning Representations, {ICLR} 2015, San Diego, CA, USA, May
  7-9, 2015, Conference Track Proceedings}, 2015.

\bibitem[Kovachki et~al.(2021)Kovachki, Li, Liu, Azizzadenesheli, Bhattacharya,
  Stuart, and Anandkumar]{kovachki21universal}
Nikola~B. Kovachki, Zongyi Li, Burigede Liu, Kamyar Azizzadenesheli, Kaushik
  Bhattacharya, Andrew~M. Stuart, and Anima Anandkumar.
\newblock Neural operator learning maps between function spaces.
\newblock \emph{CoRR}, abs/2108.08481, 2021.

\bibitem[Kurinchi{-}Vendhan et~al.(2021)Kurinchi{-}Vendhan, L{\"{u}}tjens,
  Gupta, Werner, and Newman]{kurinchi21wisosuper}
Rupa Kurinchi{-}Vendhan, Bj{\"{o}}rn L{\"{u}}tjens, Ritwik Gupta, Lucien
  Werner, and Dava Newman.
\newblock Wisosuper: Benchmarking super-resolution methods on wind and solar
  data.
\newblock \emph{Conference on Neural Information Processing Systems (NeurIPS)
  Workshop on Tackling Climate Change with Machine Learning (CCML)}, 2021.

\bibitem[Lapeyre et~al.(2019)Lapeyre, Misdariis, Cazard, Veynante, and
  Poinsot]{laypeyre19cnnparam}
Corentin~J. Lapeyre, Antony Misdariis, Nicolas Cazard, Denis Veynante, and
  Thierry Poinsot.
\newblock Training convolutional neural networks to estimate turbulent sub-grid
  scale reaction rates.
\newblock \emph{Combustion and Flame}, 203:\penalty0 255--264, 2019.

\bibitem[Lee and Carlberg(2020)]{lee20vaerom}
Kookjin Lee and Kevin~T. Carlberg.
\newblock Model reduction of dynamical systems on nonlinear manifolds using
  deep convolutional autoencoders.
\newblock \emph{Journal of Computational Physics}, 404:\penalty0 108973, 2020.
\newblock ISSN 0021-9991.

\bibitem[Li et~al.(2020)Li, Kovachki, Azizzadenesheli, Liu, Stuart,
  Bhattacharya, and Anandkumar]{li20graphoperator}
Zongyi Li, Nikola Kovachki, Kamyar Azizzadenesheli, Burigede Liu, Andrew
  Stuart, Kaushik Bhattacharya, and Anima Anandkumar.
\newblock Multipole graph neural operator for parametric partial differential
  equations.
\newblock In \emph{Advances in Neural Information Processing Systems},
  volume~33, pages 6755--6766. Curran Associates, Inc., 2020.

\bibitem[Li et~al.(2021{\natexlab{a}})Li, Kovachki, Azizzadenesheli, Liu,
  Bhattacharya, Stuart, and Anandkumar]{Li_2021}
Zongyi Li, Nikola Kovachki, Kamyar Azizzadenesheli, Burigede Liu, Kaushik
  Bhattacharya, Andrew Stuart, and Anima Anandkumar.
\newblock Fourier neural operator for parametric partial differential
  equations.
\newblock \emph{ICML}, 2021{\natexlab{a}}.

\bibitem[Li et~al.(2021{\natexlab{b}})Li, Kovachki, Azizzadenesheli, Liu,
  Bhattacharya, Stuart, and Anandkumar]{li21markov}
Zongyi Li, Nikola~B. Kovachki, Kamyar Azizzadenesheli, Burigede Liu, Kaushik
  Bhattacharya, Andrew~M. Stuart, and Anima Anandkumar.
\newblock Markov neural operators for learning chaotic systems.
\newblock \emph{CoRR}, abs/2106.06898, 2021{\natexlab{b}}.

\bibitem[Ling et~al.(2016)Ling, Kurzawski, and Templeton]{ling16rans}
Julia Ling, Andrew Kurzawski, and Jeremy Templeton.
\newblock Reynolds averaged turbulence modelling using deep neural networks
  with embedded invariance.
\newblock \emph{Journal of Fluid Mechanics}, 807:\penalty0 155–166, 2016.

\bibitem[Liu et~al.(2022)Liu, Kovachki, Li, Azizzadenesheli, Anandkumar,
  Stuart, and Bhattacharya]{liu22multiscale}
Burigede Liu, Nikola Kovachki, Zongyi Li, Kamyar Azizzadenesheli, Anima
  Anandkumar, Andrew~M. Stuart, and Kaushik Bhattacharya.
\newblock A learning-based multiscale method and its application to inelastic
  impact problems.
\newblock \emph{Journal of the Mechanics and Physics of Solids}, 158:\penalty0
  104668, 2022.
\newblock ISSN 0022-5096.
\newblock \doi{https://doi.org/10.1016/j.jmps.2021.104668}.
\newblock URL
  \url{https://www.sciencedirect.com/science/article/pii/S0022509621002982}.

\bibitem[Liu et~al.(2021)Liu, Wang, Meng, Chen, Tegmark, and
  Liu]{liu21newphysics}
Ziming Liu, Bohan Wang, Qi~Meng, Wei Chen, Max Tegmark, and Tie-Yan Liu.
\newblock Machine-learning nonconservative dynamics for new-physics detection.
\newblock \emph{Phys. Rev. E}, 104:\penalty0 055302, Nov 2021.
\newblock \doi{10.1103/PhysRevE.104.055302}.
\newblock URL \url{https://link.aps.org/doi/10.1103/PhysRevE.104.055302}.

\bibitem[Long et~al.(2018{\natexlab{a}})Long, She, and
  Mukhopadhyay]{long18hybridnet}
Yun Long, Xueyuan She, and Saibal Mukhopadhyay.
\newblock Hybridnet: Integrating model-based and data-driven learning to
  predict evolution of dynamical systems.
\newblock In \emph{Proceedings of The 2nd Conference on Robot Learning},
  volume~87 of \emph{Proceedings of Machine Learning Research}, pages 551--560.
  PMLR, 29--31 Oct 2018{\natexlab{a}}.

\bibitem[Long et~al.(2018{\natexlab{b}})Long, Lu, Ma, and Dong]{long18pdenet}
Zichao Long, Yiping Lu, Xianzhong Ma, and Bin Dong.
\newblock {PDE}-net: Learning {PDE}s from data.
\newblock In \emph{Proceedings of the 35th International Conference on Machine
  Learning}, volume~80 of \emph{Proceedings of Machine Learning Research},
  pages 3208--3216. PMLR, 10--15 Jul 2018{\natexlab{b}}.

\bibitem[Long et~al.(2019)Long, Lu, and Dong]{long19pdenet2}
Zichao Long, Yiping Lu, and Bin Dong.
\newblock Pde-net 2.0: Learning pdes from data with a numeric-symbolic hybrid
  deep network.
\newblock \emph{Journal of Computational Physics}, 399:\penalty0 108925, 2019.

\bibitem[Lorenz(2006)]{lorenz96lorenz}
Edward Lorenz.
\newblock Predictability - a problem partly solved.
\newblock In Tim Palmer and Renate Hagedorn, editors, \emph{Predictability of
  Weather and Climate}. Cambridge University Press, Cambridge, 2006.

\bibitem[Lorenz and Emanuel(1998)]{lorenz98lorenz}
Edward~N. Lorenz and Kerry~A. Emanuel.
\newblock Optimal sites for supplementary weather observations: Simulation with
  a small model.
\newblock \emph{Journal of the Atmospheric Sciences}, 55\penalty0 (3):\penalty0
  399 -- 414, 1998.

\bibitem[Lu et~al.(2021)Lu, Jin, Pang, Zhang, and Karniadakis]{lu21deeponet}
Lu~Lu, Pengzhan Jin, Guofei Pang, Zhongqiang Zhang, and George~Em Karniadakis.
\newblock Learning nonlinear operators via deeponet based on the universal
  approximation theorem of operators.
\newblock \emph{Nature Machine Intelligence}, 3:\penalty0 218--229, 2021.

\bibitem[L\"utjens et~al.(2021)L\"utjens, Crawford, Veillette, and
  Newman]{Lutjens_2021b}
Bj\"orn L\"utjens, Catherine~H. Crawford, Mark Veillette, and Dava Newman.
\newblock Pce-pinns: Physics-informed neural networks for uncertainty
  propagation in ocean modeling.
\newblock \emph{International Conference on Learning Representations (ICLR)
  Workshop on AI for Modeling Oceans and Climate Change}, May 2021.

\bibitem[L{\"u}tjens et~al.(2021)L{\"u}tjens, Crawford, Veillette, and
  Newman]{lutjens2021spectral}
Bj{\"o}rn L{\"u}tjens, Catherine~H Crawford, Mark Veillette, and Dava Newman.
\newblock Spectral {PINN}s: Fast uncertainty propagation with physics-informed
  neural networks.
\newblock In \emph{Advances in Neural Information Processing Systems (NeurIPS)
  Workshop on The Symbiosis of Deep Learning and Differential Equations
  (DLDE)}, 2021.

\bibitem[Lutter et~al.(2019)Lutter, Ritter, and Peters]{lutter2018dln}
Michael Lutter, Christian Ritter, and Jan Peters.
\newblock Deep lagrangian networks: Using physics as model prior for deep
  learning.
\newblock In \emph{International Conference on Learning Representations
  (ICLR)}, 2019.

\bibitem[McGraw and Barnes(2018)]{mcgraw18granger}
Marie~C. McGraw and Elizabeth~A. Barnes.
\newblock Memory matters: A case for granger causality in climate variability
  studies.
\newblock \emph{Journal of Climate}, 31\penalty0 (8):\penalty0 3289 -- 3300,
  2018.

\bibitem[McGuffie and Henderson-Sellers(2005)]{Mcguffie_2005}
Kendal McGuffie and Ann Henderson-Sellers.
\newblock \emph{A Climate Modeling Primer, Third Edition}.
\newblock John Wiley and Sons, Ltd., Jan. 2005.

\bibitem[Mohan et~al.(2019)Mohan, Daniel, Chertkov, and
  Livescu]{mohan19turbulencelstm}
Arvind Mohan, Don Daniel, Michael Chertkov, and Daniel Livescu.
\newblock Compressed convolutional lstm: An efficient deep learning framework
  to model high fidelity 3d turbulence, 2019.

\bibitem[Nogueira~Jr. et~al.(2021)Nogueira~Jr., Carvalho, Almeida, Codas,
  Bentivegna, and Watson]{nogueira21rom}
Alberto~C. Nogueira~Jr., Felipe C.~T. Carvalho, Jo{\~a}o Lucas~S. Almeida,
  Andres Codas, Eloisa Bentivegna, and Campbell~D. Watson.
\newblock Reservoir computing in reduced order modeling for chaotic dynamical
  systems.
\newblock In Heike Jagode, Hartwig Anzt, Hatem Ltaief, and Piotr Luszczek,
  editors, \emph{High Performance Computing}, pages 56--72, Cham, 2021.
  Springer International Publishing.

\bibitem[O'Gorman and Dwyer(2018)]{ogorman18convection}
Paul~A. O'Gorman and John~G. Dwyer.
\newblock Using machine learning to parameterize moist convection: Potential
  for modeling of climate, climate change, and extreme events.
\newblock \emph{Journal of Advances in Modeling Earth Systems}, 10\penalty0
  (10):\penalty0 2548--2563, 2018.

\bibitem[Olver(1986)]{olver86symmetry}
Peter~J. Olver.
\newblock \emph{Symmetry Groups of Differential Equations}, pages 77--185.
\newblock Springer New York, New York, NY, 1986.

\bibitem[P.(2006)]{sagaut06les}
Sagaut P.
\newblock \emph{Large Eddy Simulation for Incompressible Flows: An
  Introduction}.
\newblock Scientific Computation. Springer-Verlag Berlin Heidelberg, 3 edition,
  2006.

\bibitem[Palmer et~al.(2019)Palmer, Stevens, and Bauer]{Palmer_2019}
Tim Palmer, Bjorn Stevens, and Peter Bauer.
\newblock We need an international center for climate modeling, 2019.
\newblock URL
  \url{https://blogs.scientificamerican.com/observations/we-need-an-international-center-for-climate-modeling/}.
\newblock last accessed 04/13/20.

\bibitem[Pandey et~al.(2018)Pandey, Scheel, and Schumacher]{pandey18rbc}
Ambrish Pandey, Janet~D. Scheel, and Jörg Schumacher.
\newblock Turbulent superstructures in rayleigh-bénard convection.
\newblock \emph{Nature Communications}, 9, 2018.

\bibitem[Parish and Duraisamy(2016)]{parish16rans}
Eric~J. Parish and Karthik Duraisamy.
\newblock A paradigm for data-driven predictive modeling using field inversion
  and machine learning.
\newblock \emph{Journal of Computational Physics}, 305:\penalty0 758--774,
  2016.
\newblock ISSN 0021-9991.

\bibitem[Pathak et~al.(2018)Pathak, Hunt, Girvan, Lu, and
  Ott]{pathak18reservoircomputing}
Jaideep Pathak, Brian Hunt, Michelle Girvan, Zhixin Lu, and Edward Ott.
\newblock Model-free prediction of large spatiotemporally chaotic systems from
  data: A reservoir computing approach.
\newblock \emph{Phys. Rev. Lett.}, 120, Jan 2018.

\bibitem[Pathak et~al.(2020)Pathak, Mustafa, Kashinath, Motheau, Kurth, and
  Day]{pathak20hrres}
Jaideep Pathak, Mustafa Mustafa, Karthik Kashinath, Emmanuel Motheau, Thorsten
  Kurth, and Marcus Day.
\newblock Using machine learning to augment coarse-grid computational fluid
  dynamics simulations, 2020.

\bibitem[{Pathak} et~al.(2022){Pathak}, {Subramanian}, {Harrington}, {Raja},
  {Chattopadhyay}, {Mardani}, {Kurth}, {Hall}, {Li}, {Azizzadenesheli},
  {Hassanzadeh}, {Kashinath}, and {Anandkumar}]{pathak22fourcastnet}
Jaideep {Pathak}, Shashank {Subramanian}, Peter {Harrington}, Sanjeev {Raja},
  Ashesh {Chattopadhyay}, Morteza {Mardani}, Thorsten {Kurth}, David {Hall},
  Zongyi {Li}, Kamyar {Azizzadenesheli}, Pedram {Hassanzadeh}, Karthik
  {Kashinath}, and Animashree {Anandkumar}.
\newblock {FourCastNet: A Global Data-driven High-resolution Weather Model
  using Adaptive Fourier Neural Operators}.
\newblock February 2022.

\bibitem[Pavliotis and Stuart(2008)]{pavliotis08multiscale}
Greg Pavliotis and Andrew Stuart.
\newblock \emph{Multiscale Methods Averaging and Homogenization}, volume~53 of
  \emph{Texts in Applied Mathematics}.
\newblock Springer-Verlag New York, 1 edition, 2008.
\newblock ISBN 978-0-387-73829-1.

\bibitem[Peng et~al.(2021)Peng, Alber, Buganza~Tepole, Cannon, De, Dura-Bernal,
  Garikipati, Karniadakis, Lytton, Perdikaris, Petzold, and
  Kuhl]{peng21biomultiscale}
Grace C.~Y. Peng, Mark Alber, Adrian Buganza~Tepole, William~R. Cannon, Suvranu
  De, Savador Dura-Bernal, Krishna Garikipati, George Karniadakis, William~W.
  Lytton, Paris Perdikaris, Linda Petzold, and Ellen Kuhl.
\newblock Multiscale modeling meets machine learning: What can we learn?
\newblock \emph{Archives of Computational Methods in Engineering}, 28:\penalty0
  1017--1037, 2021.

\bibitem[{Prakash} et~al.(2021){Prakash}, {Jansen}, and {Evans}]{prakash21les}
Aviral {Prakash}, Kenneth~E. {Jansen}, and John~A. {Evans}.
\newblock {Invariant Data-Driven Subgrid Stress Modeling in the Strain-Rate
  Eigenframe for Large Eddy Simulation}.
\newblock \emph{arXiv e-prints}, June 2021.

\bibitem[Qian et~al.(2022)Qian, Kacprzyk, and van~der Schaar]{qian22dcode}
Zhaozhi Qian, Krzysztof Kacprzyk, and Mihaela van~der Schaar.
\newblock D-{CODE}: Discovering closed-form {ODE}s from observed trajectories.
\newblock In \emph{International Conference on Learning Representations}, 2022.

\bibitem[Quarteroni and Rozza(2014)]{quarteroni14roms}
Alfio Quarteroni and Gianluigi Rozza.
\newblock Reduced order methods for modeling and computational reduction, 2014.

\bibitem[Rackauckas et~al.(2020)Rackauckas, Ma, Martensen, Warner, Zubov,
  Supekar, Skinner, and Ramadhan]{Rackauckas_2020}
Christopher Rackauckas, Yingbo Ma, Julius Martensen, Collin Warner, Kirill
  Zubov, Rohit Supekar, Dominic Skinner, and Ali Ramadhan.
\newblock Universal differential equations for scientific machine learning.
\newblock \emph{ArXiv}, abs/2001.04385, 2020.

\bibitem[Raissi et~al.(2019)Raissi, Perdikaris, and Karniadakis]{raissi19pinns}
M.~Raissi, P.~Perdikaris, and G.E. Karniadakis.
\newblock Physics-informed neural networks: A deep learning framework for
  solving forward and inverse problems involving nonlinear partial differential
  equations.
\newblock \emph{Journal of Computational Physics}, 378:\penalty0 686 -- 707,
  2019.

\bibitem[Rasp(2020)]{rasp20lorenz96online}
S.~Rasp.
\newblock Coupled online learning as a way to tackle instabilities and biases
  in neural network parameterizations: general algorithms and lorenz~96 case
  study (v1.0).
\newblock \emph{Geoscientific Model Development}, 13\penalty0 (5):\penalty0
  2185--2196, 2020.

\bibitem[Rasp(2019)]{rasp19lorenztooeasy}
Stephan Rasp.
\newblock Lorenz 96 is too easy! machine learning research needs a more
  realistic toy model, July 2019.
\newblock URL \url{https://raspstephan.github.io/blog/lorenz-96-is-too-easy/}.
\newblock last accessed May 2022.

\bibitem[Rasp et~al.(2018)Rasp, Pritchard, and Gentine]{Rasp_2018}
Stephan Rasp, Michael~S. Pritchard, and Pierre Gentine.
\newblock Deep learning to represent subgrid processes in climate models.
\newblock \emph{Proceedings of the National Academy of Sciences}, 115\penalty0
  (39):\penalty0 9684--9689, 2018.

\bibitem[Rasp et~al.(2020)Rasp, Dueben, Scher, Weyn, Mouatadid, and
  Thuerey]{rasp20weatherbench}
Stephan Rasp, Peter~D. Dueben, Sebastian Scher, Jonathan~A. Weyn, Soukayna
  Mouatadid, and Nils Thuerey.
\newblock Weatherbench: A benchmark data set for data-driven weather
  forecasting.
\newblock \emph{Journal of Advances in Modeling Earth Systems}, 12\penalty0
  (11), 2020.

\bibitem[Reichstein et~al.(2019)Reichstein, Camps-Valls, Stevens, Jung,
  Denzler, Carvalhais, and Prabhat]{Reichstein_2019}
Markus Reichstein, Gustau Camps-Valls, Bjorn Stevens, Martin Jung, Joachim
  Denzler, Nuno Carvalhais, and Prabhat.
\newblock Deep learning and process understanding for data-driven earth system
  science.
\newblock \emph{Nature}, 566:\penalty0 195 -- 204, 2019.

\bibitem[Rezende and Mohamed(2015)]{rezende15normflows}
Danilo~Jimenez Rezende and Shakir Mohamed.
\newblock Variational inference with normalizing flows.
\newblock In \emph{Proceedings of the 32nd International Conference on
  International Conference on Machine Learning - Volume 37}, ICML'15, page
  1530–1538. JMLR.org, 2015.

\bibitem[Rooney-Varga et~al.(2020)Rooney-Varga, Kapmeier, Sterman, Jones,
  Putko, and Rath]{rooneyvarge20enroads}
Juliette~N. Rooney-Varga, Florian Kapmeier, John~D. Sterman, Andrew~P. Jones,
  Michele Putko, and Kenneth Rath.
\newblock The climate action simulation.
\newblock \emph{Simulation \& Gaming}, 51\penalty0 (2):\penalty0 114--140,
  2020.
\newblock URL \url{https://en-roads.climateinteractive.org/}.

\bibitem[Sirignano et~al.(2020)Sirignano, MacArt, and Freund]{sirignano20dpm}
Justin Sirignano, Jonathan~F. MacArt, and Jonathan~B. Freund.
\newblock Dpm: A deep learning pde augmentation method with application to
  large-eddy simulation.
\newblock \emph{Journal of Computational Physics}, 423, 2020.

\bibitem[Smith(2013)]{Smith_2013}
Ralph~C. Smith.
\newblock Uncertainty quantification: Theory, implementation, and applications.
\newblock In \emph{Computational science and engineering}, page 382. SIAM,
  2013.

\bibitem[SnagglebitInkArt(2022)]{Matryoshka_2022}
SnagglebitInkArt.
\newblock Sloth nesting dolls - hand painted modern russian matryoshka doll
  set, 2022.
\newblock URL
  \url{https://www.etsy.com/ie/listing/690540535/sloth-nesting-dolls-hand-painted-modern}.
\newblock last accessed 01/22.

\bibitem[Sohl-Dickstein et~al.(2015)Sohl-Dickstein, Weiss, Maheswaranathan, and
  Ganguli]{dickstein15diff}
Jascha Sohl-Dickstein, Eric Weiss, Niru Maheswaranathan, and Surya Ganguli.
\newblock Deep unsupervised learning using nonequilibrium thermodynamics.
\newblock In Francis Bach and David Blei, editors, \emph{Proceedings of the
  32nd International Conference on Machine Learning}, volume~37 of
  \emph{Proceedings of Machine Learning Research}, pages 2256--2265, Lille,
  France, 07--09 Jul 2015. PMLR.

\bibitem[Stachenfeld et~al.(2022)Stachenfeld, Fielding, Kochkov, Cranmer,
  Pfaff, Godwin, Cui, Ho, Battaglia, and
  Sanchez-Gonzalez]{stachenfeld2022learned}
Kim Stachenfeld, Drummond~Buschman Fielding, Dmitrii Kochkov, Miles Cranmer,
  Tobias Pfaff, Jonathan Godwin, Can Cui, Shirley Ho, Peter Battaglia, and
  Alvaro Sanchez-Gonzalez.
\newblock Learned simulators for turbulence.
\newblock In \emph{International Conference on Learning Representations}, 2022.

\bibitem[Stengel et~al.(2020)Stengel, Glaws, Hettinger, and
  King]{stengel20phiregan}
Karen Stengel, Andrew Glaws, Dylan Hettinger, and Ryan~N. King.
\newblock Adversarial super-resolution of climatological wind and solar data.
\newblock \emph{Proceedings of the National Academy of Sciences}, 117\penalty0
  (29):\penalty0 16805--16815, 2020.

\bibitem[Subel et~al.(2021)Subel, Chattopadhyay, Guan, and
  Hassanzadeh]{subel21multiscaleburgers}
Adam Subel, Ashesh Chattopadhyay, Yifei Guan, and Pedram Hassanzadeh.
\newblock Data-driven subgrid-scale modeling of forced burgers turbulence using
  deep learning with generalization to higher reynolds numbers via transfer
  learning.
\newblock \emph{Physics of Fluids}, 33\penalty0 (3):\penalty0 031702, 2021.

\bibitem[Thomas et~al.(2018)Thomas, Smidt, Kearnes, Yang, Li, Kohlhoff, and
  Riley]{thomas18tensorfield}
Nathaniel Thomas, Tess~E. Smidt, Steven Kearnes, Lusann Yang, Li~Li, Kai
  Kohlhoff, and Patrick Riley.
\newblock Tensor field networks: Rotation- and translation-equivariant neural
  networks for 3d point clouds.
\newblock \emph{CoRR}, abs/1802.08219, 2018.

\bibitem[Thornes et~al.(2017)Thornes, Duben, and Palmer]{thornes17lorenz96}
Tobias Thornes, Peter Duben, and Tim Palmer.
\newblock On the use of scale-dependent precision in earth system modelling.
\newblock \emph{Quarterly Journal of the Royal Meteorological Society},
  143\penalty0 (703):\penalty0 897--908, 2017.

\bibitem[Toms et~al.(2020)Toms, Barnes, and Ebert-Uphoff]{toms20interpret}
Benjamin~A. Toms, Elizabeth~A. Barnes, and Imme Ebert-Uphoff.
\newblock Physically interpretable neural networks for the geosciences:
  Applications to earth system variability.
\newblock \emph{Journal of Advances in Modeling Earth Systems}, 12\penalty0
  (9), 2020.

\bibitem[Um et~al.(2020)Um, Brand, Fei, Holl, and Thuerey]{um20diffres}
Kiwon Um, Robert Brand, Yun Fei, Philipp Holl, and Nils Thuerey.
\newblock {Solver-in-the-Loop: Learning from Differentiable Physics to Interact
  with Iterative PDE-Solvers}.
\newblock \emph{Advances in Neural Information Processing Systems}, 2020.

\bibitem[Vandal et~al.(2017)Vandal, Kodra, Ganguly, Michaelis, Nemani, and
  Ganguly]{vandal17deepsd}
Thomas Vandal, Evan Kodra, Sangram Ganguly, Andrew Michaelis, Ramakrishna
  Nemani, and Auroop~R. Ganguly.
\newblock Deepsd: Generating high resolution climate change projections through
  single image super-resolution.
\newblock In \emph{Proceedings of the 23rd ACM SIGKDD International Conference
  on Knowledge Discovery and Data Mining}, KDD '17, page 1663–1672, New York,
  NY, USA, 2017. Association for Computing Machinery.

\bibitem[Wang et~al.(2022)Wang, Yuval, and O'Gorman]{wang22nonlocal}
Peidong Wang, Janni Yuval, and Paul~A. O'Gorman.
\newblock Non-local parameterization of atmospheric subgrid processes with
  neural networks, 2022.

\bibitem[Wang et~al.(2021)Wang, Wang, and Perdikaris]{wang21deeponet}
Sifan Wang, Hanwen Wang, and Paris Perdikaris.
\newblock Learning the solution operator of parametric partial differential
  equations with physics-informed deeponets.
\newblock \emph{Science Advances}, 7\penalty0 (40), 2021.

\bibitem[Webb et~al.(2015)Webb, Lock, Bretherton, Bony, Cole, Idelkadi, Kang,
  Koshiro, Kawai, Ogura, Roehrig, Shin, Mauritsen, Sherwood, Vial, Watanabe,
  Woelfle, and Zhao]{Webb_2015}
Mark~J. Webb, Adrian~P. Lock, Christopher~S. Bretherton, Sandrine Bony,
  Jason~N.S. Cole, Abderrahmane Idelkadi, Sarah~M. Kang, Tsuyoshi Koshiro,
  Hideaki Kawai, Tomoo Ogura, Romain Roehrig, Yechul Shin, Thorsten Mauritsen,
  Steven~C. Sherwood, Jessica Vial, Masahiro Watanabe, Matthew~D. Woelfle, and
  Ming Zhao.
\newblock {The impact of parametrized convection on cloud feedback}.
\newblock \emph{Philosophical Transactions of the Royal Society A:
  Mathematical, Physical and Engineering Sciences}, 373\penalty0 (2054), nov
  2015.

\bibitem[Wilks(2005)]{wilks05lorenz96stochastic}
Daniel~S. Wilks.
\newblock Effects of stochastic parametrizations in the lorenz '96 system.
\newblock \emph{Quarterly Journal of the Royal Meteorological Society},
  131\penalty0 (606):\penalty0 389--407, 2005.

\bibitem[Willard et~al.(2022)Willard, Jia, Xu, Steinbach, and
  Kumar]{willard22scimlreview}
Jared Willard, Xiaowei Jia, Shaoming Xu, Michael Steinbach, and Vipin Kumar.
\newblock Integrating scientific knowledge with machine learning for
  engineering and environmental systems.
\newblock \emph{ACM Comput. Surv.}, jan 2022.

\bibitem[William(1991)]{schiesser91methodlines}
Schiesser William.
\newblock \emph{The Numerical Method of Lines: Integration of Partial
  Differential Equations}.
\newblock Elsevier, 1991.

\bibitem[Williams et~al.(2015)Williams, Kevrekidis, and
  Rowley]{williams15koopman}
Matthew~O. Williams, Ioannis~G. Kevrekidis, and Clarence~W. Rowley.
\newblock A data–driven approximation of the koopman operator: Extending
  dynamic mode decomposition.
\newblock \emph{Journal of Nonlinear Science}, 25:\penalty0 1432--1467, 2015.

\bibitem[Wu et~al.(2018)Wu, Xiao, and Paterson]{wu18rans}
Jin-Long Wu, Heng Xiao, and Eric Paterson.
\newblock Physics-informed machine learning approach for augmenting turbulence
  models: A comprehensive framework.
\newblock \emph{Phys. Rev. Fluids}, 3:\penalty0 074602, Jul 2018.

\bibitem[Wu et~al.(2020)Wu, Kashinath, Albert, Chirila, Prabhat, and
  Xiao]{wu20pinngans}
Jin-Long Wu, Karthik Kashinath, Adrian Albert, Dragos Chirila, Prabhat, and
  Heng Xiao.
\newblock Enforcing statistical constraints in generative adversarial networks
  for modeling chaotic dynamical systems.
\newblock \emph{Journal of Computational Physics}, 406, 2020.

\bibitem[Xie et~al.(2018)Xie, Franz, Chu, and Thuerey]{xie18tempogan}
You Xie, Erik Franz, Mengyu Chu, and Nils Thuerey.
\newblock Tempogan: A temporally coherent, volumetric gan for super-resolution
  fluid flow.
\newblock \emph{ACM Trans. Graph.}, 37\penalty0 (4), jul 2018.

\bibitem[Yazdani et~al.(2020)Yazdani, Lu, Raissi, and
  Karniadakis]{yazdani20bioparam}
Alireza Yazdani, Lu~Lu, Maziar Raissi, and George~Em Karniadakis.
\newblock Systems biology informed deep learning for inferring parameters and
  hidden dynamics.
\newblock \emph{PLOS Computational Biology}, 16\penalty0 (11):\penalty0 1--19,
  11 2020.

\bibitem[Yin et~al.(2021)Yin, Guen, Dona, de~B{\'{e}}zenac, Ayed, Thome, and
  Gallinari]{yin21aphinity}
Yuan Yin, Vincent~Le Guen, J{\'{e}}r{\'{e}}mie Dona, Emmanuel de~B{\'{e}}zenac,
  Ibrahim Ayed, Nicolas Thome, and Patrick Gallinari.
\newblock Augmenting physical models with deep networks for complex dynamics
  forecasting.
\newblock \emph{Journal of Statistical Mechanics: Theory and Experiment},
  2021\penalty0 (12):\penalty0 124012, dec 2021.
\newblock \doi{10.1088/1742-5468/ac3ae5}.
\newblock URL \url{https://doi.org/10.1088/1742-5468/ac3ae5}.

\bibitem[Yuval et~al.(2021)Yuval, O'Gorman, and Hill]{yuval21stable}
Janni Yuval, Paul~A. O'Gorman, and Chris~N. Hill.
\newblock Use of neural networks for stable, accurate and physically consistent
  parameterization of subgrid atmospheric processes with good performance at
  reduced precision.
\newblock \emph{Geophysical Research Letter}, 48:\penalty0 e2020GL091363, 2021.

\bibitem[Zeng et~al.(2021)Zeng, Wu, and Xiao]{yang21gans}
Yang Zeng, Jin-Long Wu, and Heng Xiao.
\newblock Enforcing imprecise constraints on generative adversarial networks
  for emulating physical systems.
\newblock \emph{Communications in Computational Physics}, 30\penalty0
  (3):\penalty0 635--665, 2021.

\bibitem[Zhang et~al.(2019)Zhang, Wang, and Giannakis]{zhang19power}
Liang Zhang, Gang Wang, and Georgios~B. Giannakis.
\newblock Real-time power system state estimation and forecasting via deep
  unrolled neural networks.
\newblock \emph{IEEE Transactions on Signal Processing}, 67\penalty0
  (15):\penalty0 4069--4077, 2019.

\bibitem[Zhang et~al.(2018)Zhang, Han, Wang, Car, and E]{zhang18moleculeparam}
Linfeng Zhang, Jiequn Han, Han Wang, Roberto Car, and Weinan E.
\newblock Deep potential molecular dynamics: A scalable model with the accuracy
  of quantum mechanics.
\newblock \emph{Phys. Rev. Lett.}, 120, Apr 2018.

\bibitem[Zhou et~al.(2020)Zhou, Cui, Hu, Zhang, Yang, Liu, Wang, Li, and
  Sun]{zhou20gnns}
Jie Zhou, Ganqu Cui, Shengding Hu, Zhengyan Zhang, Cheng Yang, Zhiyuan Liu,
  Lifeng Wang, Changcheng Li, and Maosong Sun.
\newblock Graph neural networks: A review of methods and applications.
\newblock \emph{AI Open}, 1:\penalty0 57--81, 2020.

\end{thebibliography}

\newpage
\appendix
\onecolumn
\section{Appendix}
\subsection{Rayleigh-Bénard Convection}\label{app:rayleigh}
\begin{figure}[h]
 \centering
 \subfigure[Ground-truth]{
  \includegraphics[width=0.9\columnwidth]{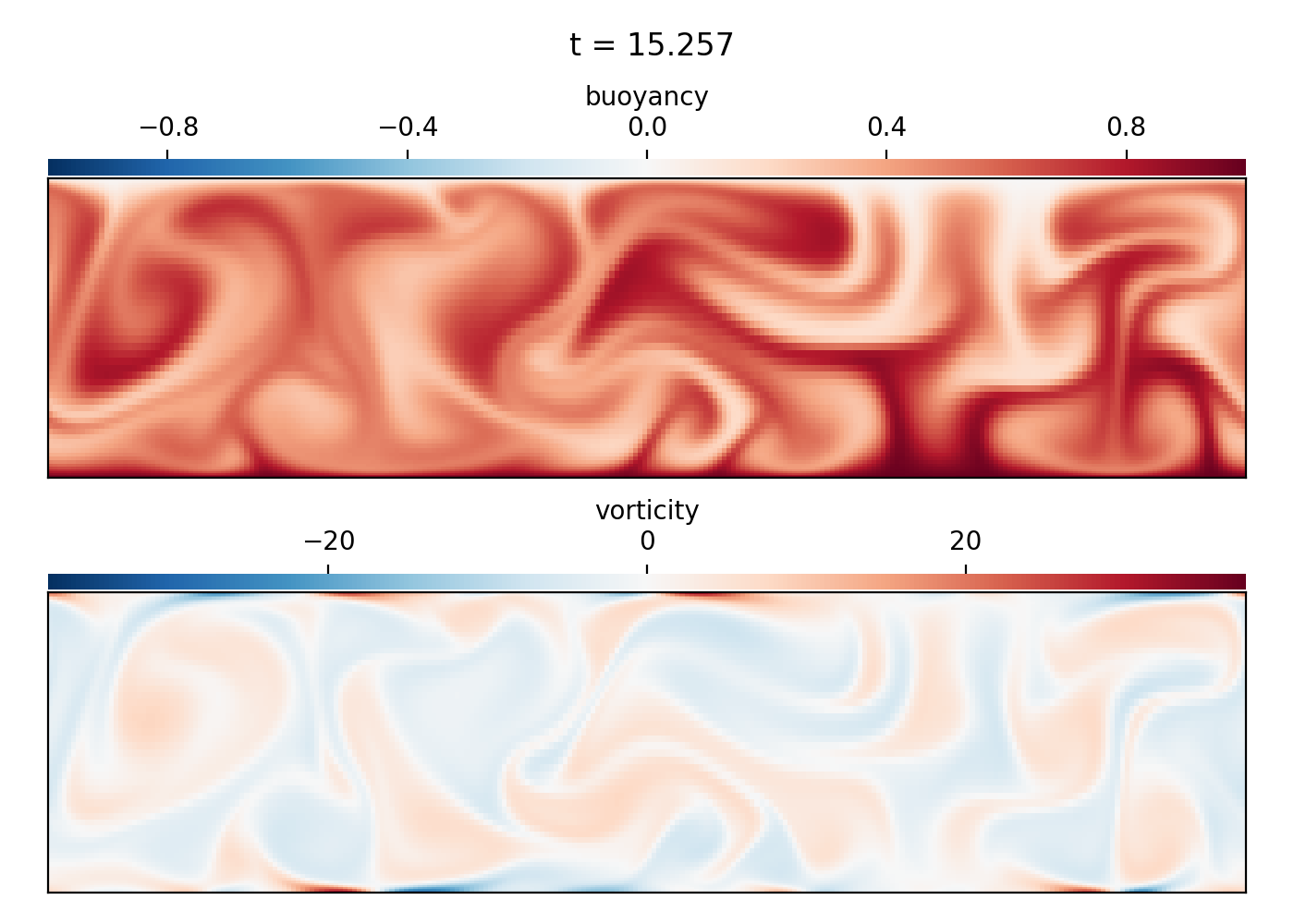}}
\caption[RayleighBenard]{\textbf{We are planning to extend MNO to Rayleigh-B\'enard Convection.} We depicted a sample plot for ground-truth training data of the 2D RBC.}
\label{fig:rayleigh}
\end{figure}

We plan to extend the multiscale neural operator to higher-dimensional systems; starting with the Rayleigh-B\'enard Convectione equations, as displayed in~\cref{fig:rayleigh}.

\subsubsection{Details and Interpretation}
Rayleigh-B\'enard Convection (RBC) is one of the simplest turbulent, chaotic, convection-dominated flows. The equation finds applications in fluid dynamics, atmospheric dynamics, radiation, phase changes, magnetic fields, and more~\citep{pandey18rbc}.

So far, we have generated a ground-truth dataset that we implemented with the 2D turbulent Rayleigh-B\'enard Convenction equations with Dedalus spectral solver~\citep{burns20dedalus} similar to~\citep{pandey18rbc}:
\begin{equation}
\begin{aligned}
\frac{\delta u}{\delta t} + u \cdot \nabla u &= \sqrt{\frac{\text{Pr}}{\text{Ra}}} \nabla^2 u - \nabla p + b \\
\frac{\delta T}{\delta t} + u \cdot \nabla T &= \frac{1}{\sqrt{Ra Pr}} \nabla^2 T \\
\nabla \cdot u = 0
\end{aligned}
\label{eq:rbc}
\end{equation}
with temperature/buoyancy, $T$, Rayleigh number, $\text{Ra}=g\alpha\Delta T H^3/(\nu\kappa)$, Prandtl number, $\text{Pr} = \nu/\kappa$, thermal expansion coefficient, $\alpha$, kinematic viscosity, $\nu$, thermal diffusivity, $\kappa=\frac{1}{\sqrt{\text{Ra}\text{Pr}}}$, acceleration due to gravity, $g$, temperature difference, $\Delta T$, unit vector, $e$, pressure, $p$, Nusselt number, $\text{Nu} = \sqrt{\frac{\text{Pr}}{\text{Ra}}}$, Reynolds number, $Re=\sqrt{ \langle \nabla^2 u\rangle_{V,t} \frac{\text{Ra}}{\text{Pr}}}$, and full volume-time average, $\langle\cdot\rangle_{V,t}$, cell length, $L_x$. The equations have been non-dimensionalized with the free-fall velocity, $U_f = \sqrt{g\alpha\Delta H}$, and cell height, $H$. In the horizontal direction, $x$, we use periodic boundary conditions and in the vertical direction, $z$, we use no-slip boundary conditions for the velocity, $u(z=0)=u(z=L_z)=0$, and fixed temperatures, $T(z=0)=L_z,\;T(z=L_z)=0$. The inital conditions are sampled randomly, $b(z, t=0) = L_z + z + z(L_z-z)\omega$, with $\omega\sim\mathcal N(0,1\times 10 ^{-3}$. 

We chose: $\text{Ra}=2\times 10^6$, $\text{Pr}=1$, $L_x=4$, $H=1$. 

\subsection{Fourier Neural Operator}\label{app:fno}

Our neural operator for learning subgrid parametrizations is based on Fourier neural operators~\citep{Li_2021}. Intuitively, the neural operator learns a parameter-to-solution mapping by learning a global convolution kernel. 
In detail, it learns the operator to transforms the current large-scale state, $\underline X(x_{0:K},t) \in \mathbb R^{K \times d_X}$ to the subgrid parametrization, $\underline{\hat f}_x(x_{0:K},t) := \underline X_{0:K} \in \mathbb R^{K\times d_X}$ with number of grid points, $K$, and input dimensionality, $d_X$, according to the following equations:

\begin{equation}
\begin{aligned}
\underline v_0 &= \underline X_{0:K} P^T + 1^{K\times 1} b_P \\
\underline v_{i+1} &= \sigma \left(\underline v_i W^T+ \int_{D_x} \kappa_\phi(x,x^\prime) v_i(x^\prime) dx^\prime \right)\\
&\approx \sigma \left(\underline v_i W^T+ 1^{n_v\times 1} b_W + \mathcal F^{-1}(R_\phi \cdot \mathcal F \underline v_i)\right)\\
\hat f_{x,0:K} &= \underline v_{n_d} Q^T+ 1^{K\times 1} b_Q
\end{aligned}
\label{eq:fno}
\end{equation}

First, MNO lifts the input via a linear transform with matrix, $P \in \mathbb R^{n_v\times d_X}$, bias, $b_P\in \mathbb R^{1\times n_v}$, vector of ones, $1^{K\times 1}$, and number of channels, $n_v$. The linear transform is local in space, i.e., the same transform is applied to each grid point. 

Second, multiple nonlinear ``Fourier layers'' are applied to the encoded/lifted state. The encoded/lifted state's, $\underline v_i\in\mathbb R^{K\times n_v}$, spatial dimension is transformed into the Fourier domain via a fast Fourier transform. We implement the FFT as a multiplication with the pre-built forward and inverse Type-I DST matrix, $\mathcal F\in \mathbb C^{k_{\max}\times K}$ and $\mathcal F^{-1}\in \mathbb C^{K \times k_{\max}}$, respectively, returning the vector, $\mathcal F\underline v_i\in \mathbb C^{k_{\max}\times n_v}$. 

The dynamics are learned via convoluting the encoded state with a weight matrix. In Fourier space, convolution is a multiplication, hence each frequency is multiplied with a complex weight matrix across the channels, such that $R\in \mathbb C^{k_{\max} \times n_v\times n_v}$.  
In parallel to the convolution with $R$, the encoded state is multiplied with the linear transform, $W \in \mathbb{R}^{n_v\times n_v}$, and bias, $b_W \in  \mathbb{R}^{1\times n_v}$. From a representation learning-perspective, the Fourier decomposition as a fast and interpretable feature extraction method that extracts smooth, periodic, and global features. The linear transform can be interpreted as residual term concisely capturing nonlinear residuals.  

So far, we have only applied linear transformations. To introduce nonlinearities, we apply a nonlinear activation function, $\sigma$, at the end of each Fourier layer. While the non-smoothness of the activation function ReLu, $\sigma(z)=\max(0,z)$, could introduce unwanted discontinuities in the solution, we choose it resulted in more accurate models than smoother activation functions such as tanh or sigmoid.

Finally, the transformed state, $v_{n_d}$, is projected back onto solution space via another linear transform, $Q \in \mathbb R^{d_X\times n_v}$, and bias, $b_Q$. 

The values of all trainable parameters, $P, R, W, Q, b_*$, are found by using a nonlinear optimization algorithm, such as stochastic gradient descent or, here, Adam~\cite{Kingma_2015}. We have used MSE between the predicted, $\hat f_x$, and ground-truth, $f_x$, subgrid parametrizations as loss. The neural operator is implemented in pytorch, but does not require an autodifferentiable PDE solver to generate training data. During implementation, we used the DFT which assumes a uniformly spaced grids, but can be exchanged with non-uniform DFTs (NUDFT) to transform non-uniform grids~\citep{dutt93nudft}.

\subsection{Multiscale Lorenz96}\label{app:lorenz96}
\subsubsection{Details and Interpretation}
The equation contains $K$ variables, $X_k\in\mathbb R$, and $JK$ small-scale variables, $Y_{j,k}\in\mathbb R$ that represent large-scale or small-scale atmospheric dynamics such as the movement of storms or formation of clouds, respectively. 
At every time-step each large-scale variable, $X_k$, influences and is influenced by $J$ small-scale variables, $Y_{0:J,k}$. The coupling could be interpreted as $X_k$ causing static instability and $Y_{j,k}$ causing drag from turbulence or latent heat fluxes from cloud formation. 
The indices $k,j$ are both interpreted as latitude, while $k\in\{0,...,K{-}1\}$ indexes boxes of latitude and $j\in\{0,...,J{-1}\}$ indexes elements inside the box. Illustrated on a 1D Earth with a circumference of $360^\circ$ that is discretized with $K=36, J=10$, one a spatial step in $k,j$ would equal $10^\circ,1^\circ$, respectively~\cite{lorenz96lorenz}; we choose $K=J=4$. A time step with $\Delta t=0.005$ would equal $36$ minutes~\cite{lorenz96lorenz}.

We choose a large forcing, $F>10$, for which the equation becomes chaotic. 
The last terms in each equation capture the interaction between small- and large-scale, $f_{x,k}=-\frac{hc}{b}\sum_{j=0}^{J}Y_{j,k}(X_k), f_y$. The scale interaction is defined by the parameters where $h=0.5$ is the coupling strength between spatial scales (with no coupling if $h$ would be zero), $b=10$ is the relative magnitude, and $c=8$ the evolution speed of $X-Y$. The linear, $-X_k$, and quadratic terms, $X_*^2$, model dissipative and advective (e.g., moving) dynamics, respectively.

The equation assumes perfect ``scale separation'' which means that small-scale variables of different grid boxes, $k$, are independent of each other at a given timestep, $Y_{j_1,k_2}(t) \bot Y_{j_2,k_1}(t)\;\forall t, j_1, j_2, k_1\neq k_2$. The separation of small- and large-scale variables can be along the same or different domain and the discretized variables would then be $\xs/\in [0, \Delta \xl/]$ or $\xs/\in [\xs/_0,\xs/_\text{end}]$, respectively.
The equation wraps around the full large- or small-scale domain by using periodic boundaries, $X_{-k}{:=}X_{K-k}$, $X_{K+k}{:=}X_k$, $Y_{-j,k}{:=}Y_{J-j,k}$, $Y_{J+j,k}{:=}Y_{j,k}$. Note that having periodic boundary conditions in the small-scale domanin allows for superparametrization, i.e., independent simulation of the small-scale dynamics~\cite{Campin_2011} and differs from the three-tier Lorenz96 where variables at the borders of the small-scale domain depend on small-scale variables of the neighbouring k~\cite{thornes17lorenz96}. 

\subsubsection{Simulation}\label{app:lorenz96_sim}
The initial conditions are sampled uniformly from a set of integers, $X(t_0) \sim U({-5, -4, ..., 5, 6})$, as a mean-zero unit-variance Gaussian $Y(t_0) \sim \mathcal N(0,1)$, and lower scale Gaussian $Z(t_0)\sim 0.05 \mathcal N(0,1)$. The train and test set contains 4k and 1k samples, respectively. Each sample is $T=1$ model time unit (MTU) or 200 (=$T/\Delta t$) time-steps long, which corresponds to $5$ Earth days ($=T/\Delta t*36\text{min}$ with $\Delta t=0.005$)~\cite{lorenz96lorenz}. 
Hence, our results test the generalization towards different initial conditions, but not robustness to extrapolation or different choices of parameters, $c,b,h,F$. The sampling starts after $T=10.$ warmup time. The dataset uses double precision. 

We solve the equation by fourth order Runge-Kutta in time with step size $\Delta t=0.005$, similar to~\citep{lorenz98lorenz}. For a PDE that is discretized with fixed time step, $\Delta t$, the ground-truth train and test data, $h_{x,0:K}(t)$, is constructed by integrating the coupled large- and small-scale dynamics.
 
Note, that the neural operator only takes in the current state of the large-scale dynamics. Hence, , i.e., it uses the full large-scale spatial domain as input, which exploits spatial correlations and learns parametrizations that are independent of the large-scale spatial discretization.

Our method can be queried for infinite time-steps into the future as it does not use time as input.

We are incorporating the prior knowledge from physics by calculating the large-scale dynamics, $dX_{LS,0:K}$. Note that the small-scale physics do not need to be known. Hence, MNO could be applied to any fixed time-step dataset for which an approximate model is known. 

\subsection{Appendix to Illustration of MNO via multiscale Lorenz96}\label{app:lorenz96_mno}
The other large-scale (LS) and fine-scale (FS) terms are 
\begin{equation}
\begin{aligned}
&\text{filtered FS dynamics, }\overline{\mathcal N(u)}(x) = 
\begin{cases}
  \frac{\delta X_k}{\delta t}  &\text{if } x = k(J+1)\;\forall k\in\{0,\dots,K\} \\
  0 &\text{otherwise}
\end{cases}\\
&\text{LS dynamics, }\mathcal N(\bar u)(x) = 
\begin{cases}
  \frac{\delta \bar X_k}{\delta t}  &\text{if } x = k(J+1)\;\forall k\in\{0,\dots,K\} \\
  0 &\text{otherwise}
\end{cases}\\
&\text{with abbreviation, }\frac{\delta \bar X_k}{\delta t} :=  X_{k-1}( X_{k+1}- X_{k-2})- X_k + F\\
&\text{LS state, }\bar u(x) = \mathcal G\ast u(x) = [X_0, 0, ..., 0, X_1, 0,..., X_K]
\end{aligned}
\label{eq:multiscale_lorenz96_app}
\end{equation}

\subsection{Appendix to Results}\label{app:results}
\subsubsection{Accuracy}\label{app:accuracy}
\begin{figure}[t]
 \centering
  \subfigure{
  \includegraphics[width=0.7\columnwidth]{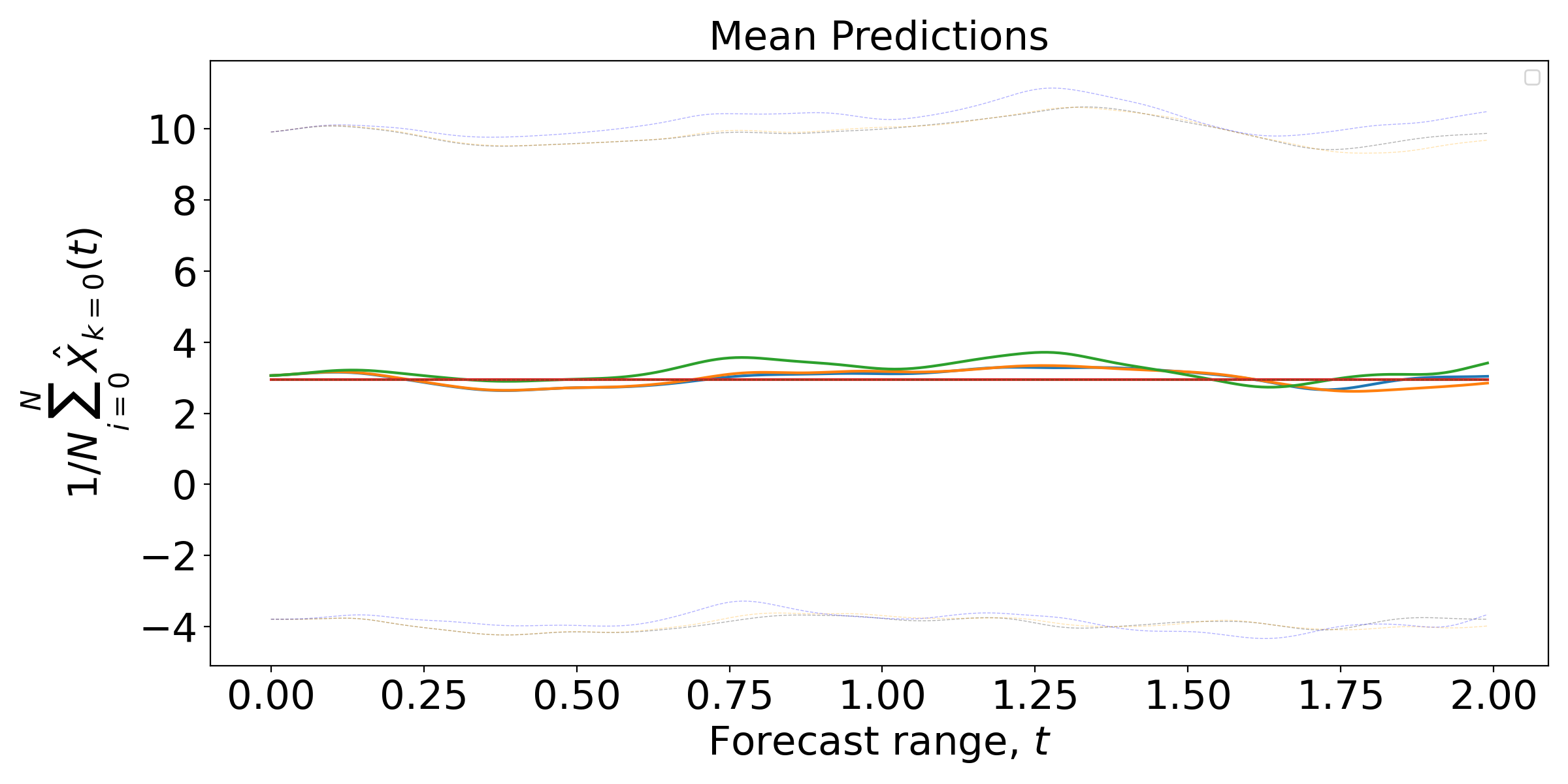}}
\caption[predictions_avg]{\textbf{Mean accuracy.} MNO (orange) most accuracy forecasts the mean (solid) and standard deviation (dotted) of the ground-truth DNS (blue) in comparison to ML-based parametrizations (green) and climatology (red).}.
\label{fig:mean_accuracy} 
\end{figure}

\Cref{fig:mean_accuracy} shows that the predicted mean and standard deviation of MNO (orange) closely follows the ground-truth (blue). The ML-based parametrization (green) follows the ground-truth only for a few time steps (until $\sim t=0.125$). The climatology (red) depicts the average prediction in the training dataset. 

\subsubsection{Model configuration}\label{app:hyperparam}
\paragraph{Multiscale Lorenz96: MNO}
As hyperparameters we chose the number of channels, $n_v=64$, number of retained modes, $k_{\max}=3$, number of Fourier layers, $n_d=3$, and no batch norm layer. The time-series modeling task uses a history of only one time step to learn chaotic dynamics~\citep{li21markov}. We are using ADAM optimizer with learning rate, $\lambda=0.001$, step size, $20$, number of epochs, $n_e=2$, and an exponential learning rate scheduler with gamma, $\gamma=0.9$~\citep{Kingma_2015}. Training took $1:50$min on a single core Intel i7-7500U CPU@2.70GHz.   
\paragraph{Multiscale Lorenz96: ML-based parametrization}
The ML-based parametrizations uses a ResNet with $n_\text{layers}=2$ residual layers that contain a fully connected network with $n_\text{units} = 32$ units. The model is optimized with Adam~\citep{Kingma_2015} with learning rate $0.01$, $\beta=(0.9, 0.999)$, $\epsilon=1*10^{-8}$, trained for $20 n_\text{epochs}=20$.
\paragraph{Multiscale Lorenz96: Traditional parametrization}
The traditional parametrization uses least-squares to find the best linear fit. The weight matrix is computed with $A = (X^T X)^{-1} X^T Y$, where $X$ and $Y$ are the concatenation of input large-scale features and target parametrizations, respectively. Inference is conducted with $\hat y = A x$.

\subsection{Neural networks vs. neural operators}
Most work in physics-informed machine learning relies on fully-connected neural networks (FCNNs) or convolutional neural networks~\citep{Karniadakis_2021}. FCNNs however are mappings between finite-dimensional spaces and learn mappings for single equation instances rather than learning the PDE solver. In our case FCNNs only learn mappings on fixed spatial grids. We leverage the recently formulated neural operators to extend the formulation to arbitrary grids. The key distinction is that the FCNN learns a parameter-dependent set of weights, $\Phi_{\as/}$, that has to be retrained for every new parameter setting. The neural operator is a learned function mapping with parameter-independent weights, $\Theta$, that takes parameter settings as input and returns a function over the spatial domain, $G_\Theta(\as/)$. In comparison, the forcing term is approximated by an FCNN as $\hat f_{x,\Phi}(\xlk/;\as/) = g_{\Phi_{\as/}}(\xlk/)$ and by a neural operator as $\hat f_{x,\Theta}(\xlk/;\as/) = G_\Theta({\as/})(\xlk/)$. The mappings are given by:
\begin{equation}
\begin{aligned}
\text{FCNN: } g_{\Phi_{\as/}}:\;D_x &\rightarrow \mathbb{R}^{d_{\soll/}}, \\
\text{NO: } G_\Theta:\;H_{\as/}(D_x;\mathbb{R}^{d_{\as/}})&\rightarrow H_{\soll/}(D_x;\mathbb{R}^{d_{\soll/}}).
\end{aligned}
\label{eq:neural_op}
\end{equation}
$H_{\as/}$ is a function space (Banach) of PDE parameter functions, $\as/$, that map the spatial domain, $D_{y}$, onto $d_{\as/}$ dimensional parameters, such as ICs, BCs, parameters, or forcing terms. 
$H_{\soll/}$ is the function space of residuals that map the spatial domain, $D_x$, onto the space of $d_{\soll/}$-dimensional residuals, $\mathbb R^{d_{\soll/}}$.

\end{document}